\documentclass[runningheads]{llncs}

\usepackage{eccv}

\usepackage{eccvabbrv}

\usepackage{graphicx}
\usepackage{booktabs}
\usepackage{amsmath}
\usepackage{graphicx}    %
\usepackage{pifont}
\usepackage{comment}
\usepackage{multirow}
\usepackage{caption}
\definecolor{light-red}{rgb}{1.,.4,.4}
\newcommand{\xmark}{\ding{55}}%
\newcommand{\revised}[1]{{\textcolor{black}{#1}}}

\usepackage[accsupp]{axessibility}  

\usepackage{hyperref}

\usepackage{orcidlink}

\begin{document}

\title{Appearance-Based Refinement for \\ Object-Centric Motion Segmentation} 


\author{Junyu Xie\inst{1}\orcidlink{0009-0002-1123-493X}\and
Weidi Xie\inst{1,2}\orcidlink{0009-0002-8609-6826}
\and Andrew Zisserman\inst{1}\orcidlink{0000-0002-8945-8573}}

\authorrunning{J.~Xie et al.}

\institute{Visual Geometry Group, Dept.\ of Engineering Science, University of Oxford\and
CMIC, Shanghai Jiao Tong University \\
\email{\{jyx,weidi,az\}@robots.ox.ac.uk}
\url{https://www.robots.ox.ac.uk/vgg/research/appear-refine/} 
}

\maketitle
\begin{abstract}
The goal of this paper is to discover, segment, and track independently moving objects in complex visual scenes. Previous approaches have explored the use of optical flow for motion segmentation, leading to imperfect predictions due to partial motion, background distraction, and object articulations and interactions. To address this issue, we introduce an appearance-based refinement method that leverages temporal consistency in video streams to correct inaccurate flow-based proposals. Our approach involves a sequence-level selection mechanism that identifies accurate flow-predicted masks as exemplars, and an object-centric architecture that refines problematic masks based on exemplar information. The model is pre-trained on synthetic data and then adapted to real-world videos in a self-supervised manner, eliminating the need for human annotations. Its performance is evaluated on multiple video segmentation benchmarks, including DAVIS, YouTubeVOS, SegTrackv2, and FBMS-59. We achieve competitive performance on single-object segmentation, while significantly outperforming existing models on the more challenging problem of multi-object segmentation. Finally, we investigate the benefits of using our model as a prompt for the per-frame Segment Anything Model. 
\end{abstract}

\section{Introduction}

\begin{figure}[htb]
\centering
\includegraphics[width=0.83\textwidth]{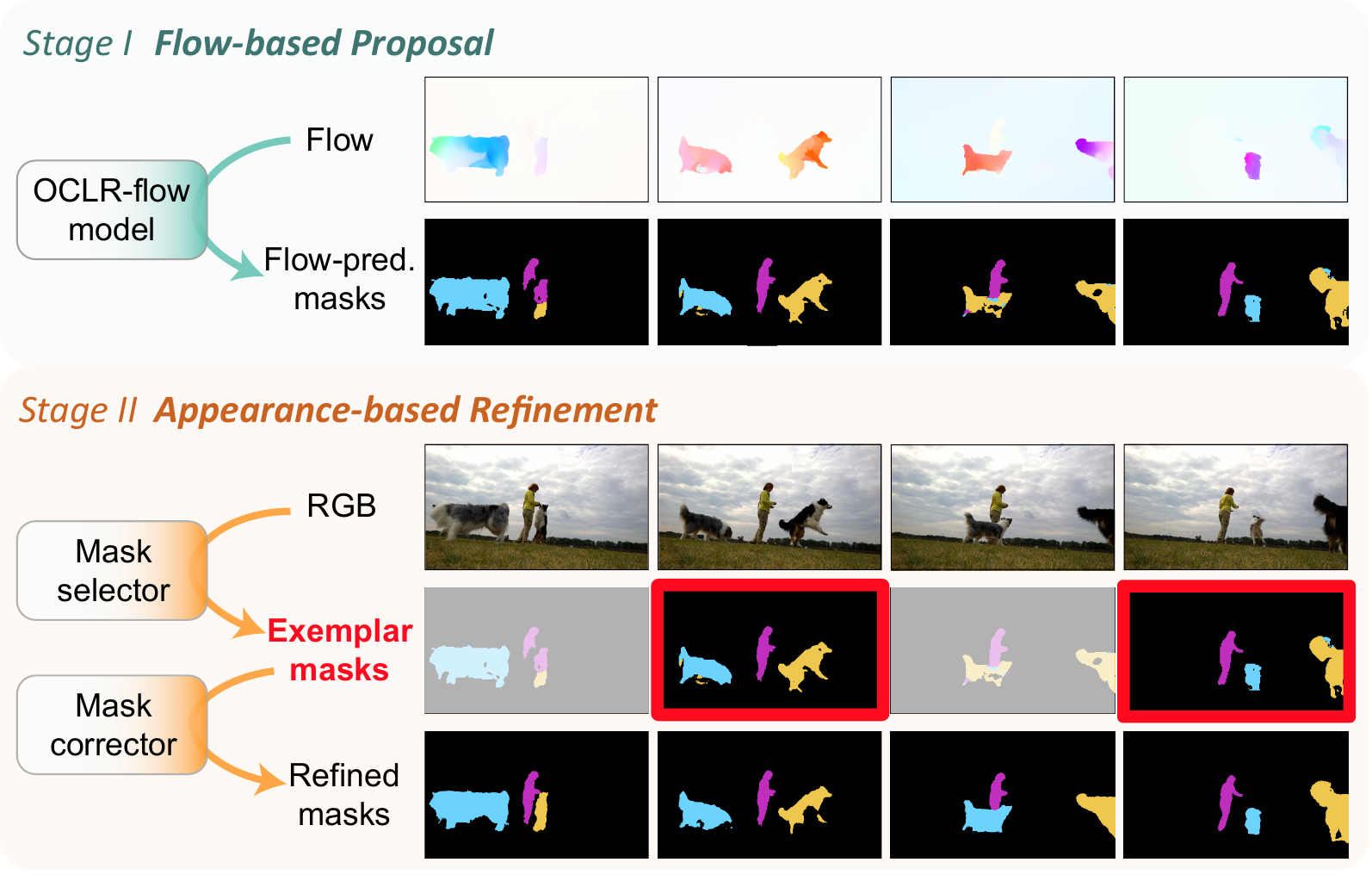}
    \caption{
    \textbf{Overview of our multi-object motion segmentation method.} The method starts with proposing object masks based on optical flow inputs ({\em i.e.,~}\textbf{flow-based proposals}), followed by an \textbf{appearance-based refinement} of flow-predicted masks. The latter stage relies on a selection-correction procedure, where high-quality exemplar masks are selected to guide the correction of other masks.  Mask selection involves picking high-quality proposal masks based on both temporal coherence across mask shapes and semantic consistency with object appearances. Mask correction involves using the selected masks (the exemplars) to guide an appearance-based mask correction process in an object-centric model. The success of the method can be seen in the bottom row where correct segmentation masks are predicted, overcoming the deficiencies of the flow-predicted masks in Stage 1. }
    \label{fig:overview}
\end{figure}
Given a video of a complex visual scene, 
we consider the task of discovering, segmenting and tracking the independently moving objects. 
Recent approaches for this task have built on two key advantages of optical flow inputs~\cite{yang_loquercio_2019, Yang21a, Lamdouar21, Xie22}. {\em First}, segmentation masks can be directly obtained from discontinuities in the flow field; 
{\em Second}, optical flow eliminates the domain gap between synthetic and real videos, making it possible to train models on synthetic sequences and apply them directly to real sequences without incurring a significant Sim2Real penalty.

While optical flow has these advantages, 
there are also some drawbacks to relying solely on flow. 
{\em First}, multi-object motion patterns in real-world videos {\em are complex} -- they can include objects that have no motion relative to the camera in some frames (and so no discontinuities), and objects can be articulated, can interact with other objects, and be partly or entirely occluded. Additionally, the background can have distractor motions, which can lead to ambiguous groupings and segmentations. 
{\em Second}, unlike the appearance stream where there is a consistency of texture and colour over the video, video constancy is not guaranteed in flow fields. This makes it challenging to establish long-term correlations and build object permanence, which can be crucial in tracking objects over time.

In this paper, to address the drawbacks of flow, we incorporate temporally-consistent appearance information by using a \textit{two-stream} approach~\cite{Simonyan14b, Feichtenhofer16, faktor2014videonlc} for the task of object discovery, segmentation, and tracking in videos. The motion stream (flow) is used for object discovery, while the appearance stream is used for maintaining consistency in segmentation and tracking, leveraging long-term correlations that arise from the persistence of appearance.

Our approach is depicted in Figure~\ref{fig:overview} and consists of two stages. 
In the first stage, we use an off-the-shelf flow-based motion segmentation model, OCLR~\cite{Xie22}.
As noted above, the flow can be used to discover moving objects (and has little Sim2Real gap), but this is as far as OCLR-flow can go -- the model generates proposals for segmentation masks in{\em~each} frame, but these vary in quality due to indistinct and partially occluded object motion. Appearance can give stronger segmentations, yet faces a significant Sim2Real gap. The novelty in this paper lies in our solution to this problem -- ``How can appearance information be used to improve flow-based segmentation?'' -- with three distinct innovations that form the second stage of the approach: (i) we propose a sequence-level mask {\em selection} mechanism, with DINO features~\cite{caron2021emerging} employed for establishing temporal consistency; 
(ii) the selected masks (that we treat as exemplars) support mask {\em correction}, where we employ a combination of direct mask supervision and motion-initialised semantic reconstruction for each frame; (iii) a self-supervised adaptation technique is introduced to effectively address the Sim2Real gap 
(that has some similarities with~\cite{wang2023testtime}). 
This selection-correction procedure enables accurate mask predictions under complex motion scenarios, including partial object motion, background distraction, object articulation, and interaction, {\em etc.}

\noindent In summary, we make the following five contributions:  
\begin{enumerate}
    \item We introduce \textbf{a sequence-level selection mechanism} designed to globally identify exemplar object masks from flow-based segmentation predictions over a long temporal range. 
    
    \item We propose \textbf{an object-centric mask correction network} that incorporates both local ({\em i.e.,}~short-term) and global ({\em i.e.,}~exemplar) information into motion-initialised object queries. Additionally, we investigate a unique loss formulation that combines direct mask supervision with motion-aware semantic reconstruction.
    
    \item We follow \textbf{a Sim2Real training procedure} that relies on \textit{zero human annotation}: The framework is synthetically pre-trained, followed by a self-supervised training scheme designed for simple yet effective domain adaptation to real-world video sequences.
    
    \item Our model is evaluated on multiple motion segmentation benchmarks, including DAVIS, YouTubeVOS, SegTrackv2, and FBMS-59, demonstrating competitive performance on single-object segmentation, while significantly outperforming existing approaches under challenging multi-object scenarios.  
    
    \item In further exploration, we use our video-level predictions as prompts for the Segment Anything Model (SAM)~\cite{kirillov2023segany}. This reveals that the object-centric consistency surpasses direct per-frame segmentation, also highlighting the complementary improvements that SAM could bring to our results. 
\end{enumerate}

\section{Related Work}
\label{sec:related}

\par{\noindent \textbf{Motion Segmentation}} 
aims to discover and segment objects from motion cues in videos. Many prior works~\cite{Bro10c,6126418,7410731,8953201} approach this problem by clustering pixels based on similar motion patterns; Other studies~\cite{Bideau16, 8578158, Mahendran_2018_ECCV_Workshops, arxiv.2201.02074} apply different motion models to explain the composition of optical flows and yield segmentation predictions. 
Innovations in model architectures~\cite{10.1007/s11263-018-1122-2, Tokmakov_2017_CVPR, Dave2019TowardsSA} and training strategies~\cite{yang_loquercio_2019} also lead to significant progress in motion segmentation accuracy. Yang et al.~\cite{Yang21a} apply the Slot Attention mechanism~\cite{locatello2020object} to reconstruct optical flows based on object slots. Lamdouar et al.~\cite{Lamdouar21} and Xie et al.~\cite{Xie22} explore a Sim2Real protocol to train a Transformer-based architecture that aggregates flow information in a temporally consistent manner. 

Recent research~\cite{faktor2014videonlc, Bao_2022_CVPR, ye2022sprites} has also focused on incorporating appearance information with motion cues. 
Choudhury et al.~\cite{Choudhury22} and Ponimatkin et al.~\cite{10030403}  leverage spectral clustering on appearance features to improve segmentation accuracy. DyStaB~\cite{Yang_2021_CVPR} trains dynamic and static models via an iterative bootstrapping strategy. Our method follows a similar two-stream idea and targets a more challenging multi-object motion scenario.

\par{\noindent \textbf{Video Object Segmentation (VOS)}} focuses on obtaining pixel-level masks of one or multiple salient objects in videos. As defined by several benchmark datasets~\cite{Perazzi16,Caelles_arXiv_2019},  video object segmentation tasks can be further divided into two main protocols, namely semi-supervised VOS~\cite{Vondrick18,Lai19,Lai20,Miao2022mamp,CVPR2019_CycleTime,jabri2020walk,caron2021emerging} and unsupervised VOS~\cite{Lu_2019_CVPR,Ventura_2019_CVPR,cho2019key,Li_2018_CVPR,luiten2020unovost}. The former provides first-frame ground-truth annotation and aims to recover segmentation masks for the remaining frames, while the latter automatically discovers and tracks salient objects from raw videos. Albeit referred to as ``unsupervised'', some previous studies rely on models that are pre-trained using manual labels, whereas our proposed method does not rely on any human annotations.

\par{\noindent \textbf{Sequence-Level Frame Selection}}
aims to pick the most informative frames from a video sequence, 
enhancing efficiency in downstream applications. This idea is widely explored across various computer vision domains, including action recognition~\cite{Raptis_2013_CVPR,Zhu_2016_CVPR,playfair}, video retrieval~\cite{2022_centerclip,Buch_2022_CVPR}, {\em etc.} 
In video segmentation, BubbleNet~\cite{GrCoCVPR19} employs this mechanism to select a single RGB frame that initiates mask propagation for semi-supervised VOS, while Yin et al.~\cite{Yin_2021_CVPR} extends this idea to interactive VOS. Wu et al.~\cite{wu2020memory} introduce a Memory Selection Network for memory-based mask propagation. Recent studies~\cite{Kuznetsova_2021_WACV,Delatolas_2024_WACV,Bekuzarov_2023_ICCV} also utilise temporal selection to propose key frames for efficient segmentation annotations. Unlike prior works that focus on RGB frame selections via confidence score estimations, our selection network regards error map predictions as a proxy task to guide the exemplar \textbf{mask} selection for \textbf{each object}.

\par{\noindent \textbf{Two-Stream Approaches.}}
The idea originates from the two-stream hypothesis in neuroscience~\cite{Schneider1962, GOODALE199220, Norman2022}, which suggests that the human visual cortex processes high temporal frequency motion information in the dorsal stream, and slower temporal frequency appearance information in the ventral stream. This two-stream (motion \& appearance) approach has been successfully applied to tasks such as action recognition~\cite{Simonyan14b, Feichtenhofer16, ma2018ts, Feichtenhofer_2019_ICCV} and VOS~\cite{faktor2014videonlc, Yang19, zhou20, Yang_2021_CVPR, BATMAN, Miao2022mamp, liu2023instmove}.

\par{\noindent \textbf{Object-Centric Representation}} decomposes complex scenes as objects (and the background components) based on reconstruction cues. Recently, unsupervised object-centric representation learning for both static scenes~\cite{NIPS2016_01eee509,NIPS2016_52947e0a,Greff19,Burgess19,Engelcke20, bear2020learning} and videos~\cite{He_2019_CVPR,NEURIPS2018_7417744a,arxiv.2106.03849} has drawn increasing research interest. 
Slot Attention~\cite{locatello2020object} proposes a fixed number of slots that compete for explaining parts of the input image through an iterative binding process. This idea is later widely adopted in video object representation learning~\cite{OP3_19, Jiang2020SCALOR, Weis2021, Kipf22, elsayed2022savipp}. 

Early unsupervised object-centric learning research mainly focuses on synthetic datasets, while more recent works achieve object segmentation and tracking on real-world videos. Fan et al.~\cite{fan2023unsupervised} and Aydemir et al.~\cite{Aydemir23} achieve an object-centric video representation by employing reconstruction of the self-supervised visual features, thereby extending the method of Seitzer et al.~\cite{seitzer2023bridging} from real-world images to videos.

\section{Appearance-Based Selection and Correction}
\label{sec:method}

Given an off-the-shelf multi-object motion segmentation model, 
we aim to refine the imperfect predictions from flow-only signals, 
by exploiting the temporal consistency in video appearance streams.

\subsection{Problem Scenario}
\label{sec:prob}

We begin with a video sequence of $T$ frames and their corresponding optical flow, denoted as $\mathcal{V}_{\text{rgb}} = \{I_1, I_2, \dots, I_T\}$ and $\mathcal{V}_{\text{flow}} = \{F_1, F_2, \dots, F_T\}$, respectively. 
We adopt OCLR-flow~\cite{Xie22} as an off-the-shelf multi-object discovery model, that takes in optical flow as input and outputs multi-object segmentation across different frames. The model is defined as follows:
\begin{equation}
\{\hat{\mathcal{M}}_1^k, \dots, \hat{\mathcal{M}}_{T}^k\}_{k=1}^{K}= \Phi_{\text{OCLR-flow}}(\mathcal{V}_{\text{flow}})
\end{equation}
$\hat{\mathcal{M}}_i^k \in \mathbb{R}^{H \times W \times 1}$ represents the flow-predicted segmentation mask for the $k^{\text{th}}$ object in frame $i$, 
where $H$ and $W$ denote the height and width of the output mask. 
The masks with the same superscript either correspond to a unique object along the video sequence ({\em i.e.,~}tracking one object), or are empty.

Our goal is to refine motion segmentation by leveraging the video appearance stream,
specifically, it involves selecting a set of high-quality exemplar segmentation masks and refining them by exploiting the temporal consistency in the appearance stream. 
To achieve this, we use the flow-predicted masks as a given prior and apply the following model:

\begin{equation}
\{\tilde{\mathcal{M}}_1^k, \dots, \tilde{\mathcal{M}}_{T}^k\}_{k=1}^K = \Phi_{\text{corrector}} \circ \Phi_{\text{selector}} (\mathcal{V}_{\text{RGB}}; \{\hat{\mathcal{M}}_1^k, \dots, \hat{\mathcal{M}}_{T}^k\}_{k=1}^K)
\end{equation}
where $\tilde{\mathcal{M}}_i^k \in \mathbb{R}^{H \times W \times 1}$ denotes the refined mask output for the $k^{\text{th}}$ object in frame $i$. The selection network ($\Phi_{\text{selector}}$) is applied to identify high-quality exemplar masks, followed by a correction network ($\Phi_{\text{corrector}}$) to conduct further mask refinements.

\subsection{Appearance-Based Mask Selection}
\label{sec:selection}

\begin{figure}[t!]
    \centering
    \includegraphics[width=0.98\textwidth]{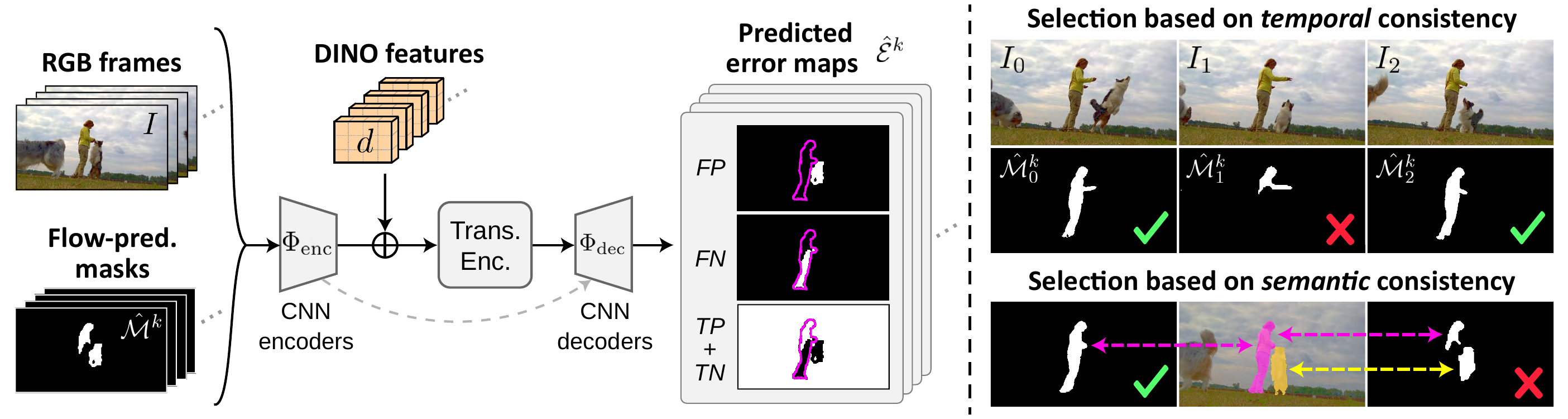}
    \caption{\textbf{Appearance-based mask selection.}
    \textbf{Left:} The mask selector takes in RGB frames, DINO features, and flow-predicted masks, generating three-channel error maps for each object (in this case the standing person). The error maps are used to select the exemplar masks for each object.
    The false positive (FP) channel highlights the over-segmented areas (the dog), while the false negative (FN) channel predicts the missing masks (the lower body of the person). 
    The remaining channel encompasses a combination of true positive and true negative (TP + TN) regions, where the flow-predicted masks align accurately with the groundtruth, as indicated by white regions. 
    The groundtruth annotation boundaries (purple contours) are provided for illustration purposes only.  \textbf{Right:} Two potential cues underlying the mask selection process:  (i) temporal consistency of mask shapes across consecutive frames, and 
    and (ii) the consistency of semantic content
    -- the mask (including all its parts) should only correspond to one semantic class. 
    }
    \label{fig:rgb_selector}
    \vspace{-0.2cm}
\end{figure}

We propose a mask selector that takes the flow-predicted masks and RGB frames as inputs, to identify a set of high-quality exemplar masks \textbf{independently} for each object $k$. Formally, 
\begin{equation}
  (\hat{\mathcal{E}}_1^k,  \dots, \hat{\mathcal{E}}_T^k)  = \Phi_{\text{selector}}(\mathcal{V}_{\text{RGB}}; \hat{\mathcal{M}}_1^k, \dots, \hat{\mathcal{M}}_{T}^k)
  {\;\longrightarrow (\text{\small$\hat{\mathcal{M}_e}$}_1^k, \dots, \text{\small$\hat{\mathcal{M}_e}$}_{t}^k)}
\end{equation}
where $\hat{\mathcal{E}}_1^k, \dots, \hat{\mathcal{E}}_T^k$ are predicted error maps,   
from which a set of $t$ \textbf{exemplar masks} {\small $\hat{\mathcal{M}_e}_1^k, \dots, \hat{\mathcal{M}_e}_{t}^k$} can be determined for \textbf{each} object by comparing the size of predicted erroneous regions.

By considering multiple frames and their associated {flow-based mask proposals}, the selection procedure is able to employ two potential cues, as illustrated in Figure~\ref{fig:rgb_selector}:
(i) the temporal consistency between neighbouring masks --
objects undergo smooth and continuous changes in position and shape, and
the selector can  filter out the frames with empty or incomplete masks due to stationary objects ({{\em i.e.,~}}no flow) or abrupt occlusions;
(ii) the consistency of semantic content within each RGB frame -- as a high-quality object mask normally corresponds to only one semantic class and this does not switch throughout the video.

\vspace{0.1cm}
\par\noindent {\bf Architecture. }
Figure~\ref{fig:rgb_selector} illustrates a simple architecture for the mask selector that considers a joint input of RGB frames and flow-predicted masks at all time steps. 
To incorporate semantic priors, {visual features from the frozen DINO ViT~\cite{caron2021emerging} are concatenated channel-wise} with the CNN-encoded features in the model bottleneck. 
These DINO features have been shown to be effective for encoding frame-to-frame correspondence~\cite{LOST, Melaskyriazi22, wang2022tokencut, shin2022selfmask}, and thus enable the transformer to use the two potential cues for the selection. Full-resolution error maps are obtained from a UNet-like upsampling process, where the original RGB frames help by providing sharp boundaries.

 \vspace{0.1cm}
\noindent {\bf Training Objective. }
In contrast to prior approaches~\cite{GrCoCVPR19,Yin_2021_CVPR} that directly compute confidence scores, we task the selector to predict an error map, {\em i.e.,~}the difference map between the input flow-predicted mask and the groundtruth segmentation, as an intermediate result.
Specifically, for each frame $i$, the prediction $\hat{\mathcal{E}}_i \in \mathbb{R}^{H \times W \times 3}$ consists of three channels corresponding to the false positive (FP), false negative (FN) and a combination of true positive and true negative (TP + TN) regions. 
We train the mask selector by applying a binary cross-entropy loss on each of the predicted regions:
\begin{equation}
    \label{loss:errormap}
    \mathcal{L}_{\text{selector}} = \frac{1}{T}\sum_{i=1}^T\left( \mathcal{L}_{\text{bce}}(\hat{\mathcal{E}}_i, \mathcal{E}_i)  \right)
\end{equation}
The groundtruth error map, $\mathcal{E}_i$, is derived from a synthetic video dataset (as will be detailed in Sec.~\ref{sec:train}), which serves as the training data for the model. By comparing the size of erroneous regions ({\em i.e.,~}FP and FN regions) across the sequence, we can compute a score that indicates the mask quality for each frame, and subsequently keep $t$ exemplar masks with high scores for each object.

\subsection{Appearance-Based Mask Correction}
\label{sec:correction}

\begin{figure*}[t]
    \centering
    \includegraphics[width=0.88\textwidth]{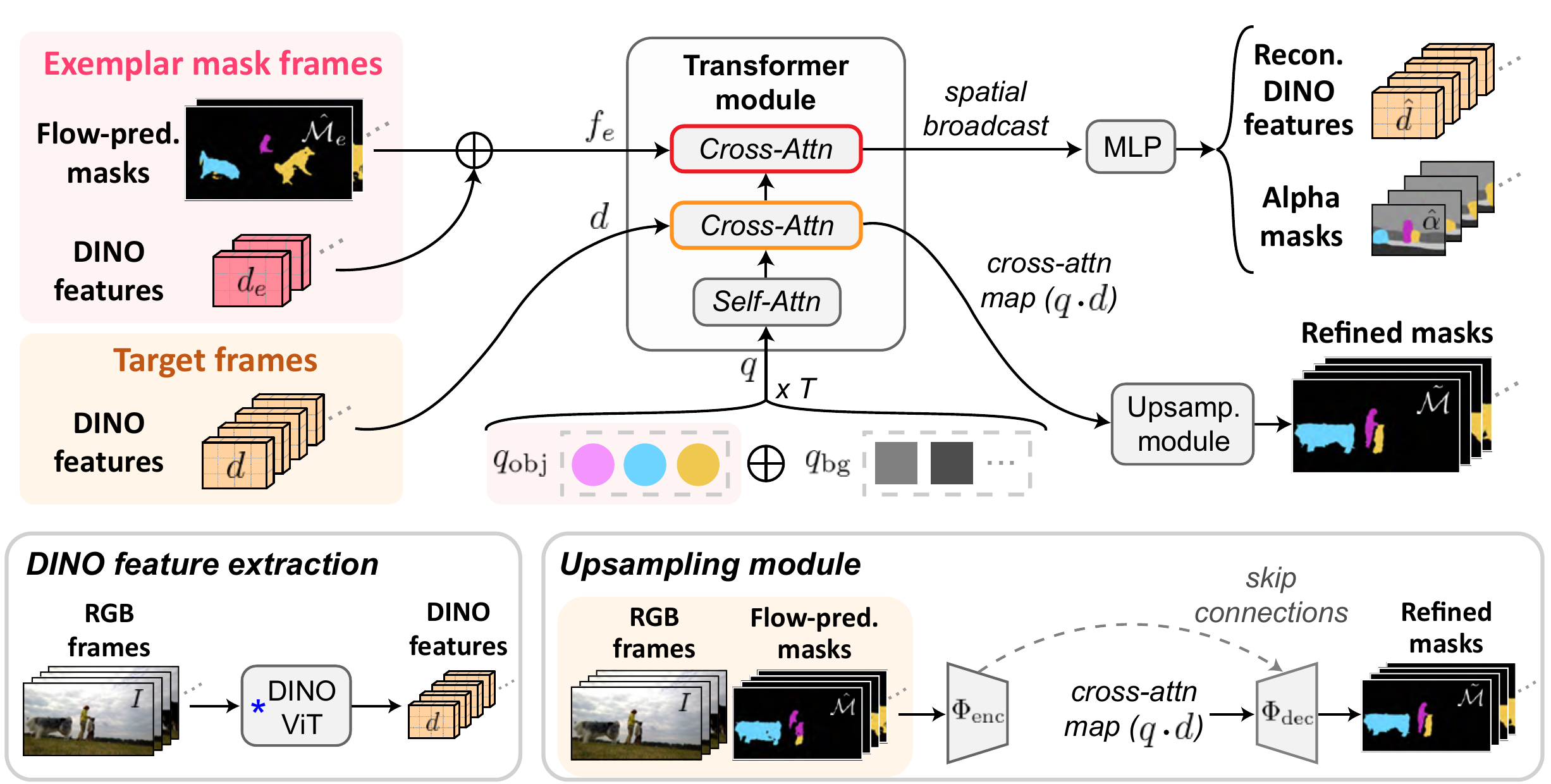}
    \caption{\textbf{Appearance-based mask corrector.} 
    The object-centric architecture first utilises 
    the exemplar mask information to initialise independent object queries. The transformer decoder module then refines the query vectors with two cross-attention layers using 
    dense target frame features for one, and exemplar information for the other.
    There are two major outputs: (i) the refined queries are spatially broadcast and projected to recover  DINO features in target frames, and (ii) the cross-attention maps between queries and target frame DINO features are extracted, followed by a CNN-based upsampling module to yield final refined masks.
    Note, ``Exemplar mask frames'' refer to frames from which the exemplar masks are selected. For illustration, we present exemplar masks for different objects in the same frame. However, in practice, different objects may have distinct sets of exemplar mask frames.  
    }
    \label{fig:rgb_corrector}
\end{figure*}

Given the high-quality exemplars from the selector, the object-centric corrector aims to refine the flow-predicted object masks across $T$ contiguous \textbf{target frames} of the sequence. 
This correction process essentially uses strong framewise correlations to facilitate generalisation of the sparse exemplar mask information to other (target) frames. Additionally, it recovers the fine-grained details by leveraging the input appearance information and the sharp boundaries in RGB frames and flow-predicted masks. The correction can be written as:
\begin{equation}
\{\tilde{\mathcal{M}}_1^k, \dots, \tilde{\mathcal{M}}_{T}^k\}_{k=1}^K  = \Phi_{\text{corrector}} (\mathcal{V}_{\text{RGB}}; \{\hat{\mathcal{M}}_1^k, \dots, \hat{\mathcal{M}}_{T}^k; {\text{\small$\hat{\mathcal{M}_e}$}_1^k, \dots, \text{\small$\hat{\mathcal{M}_e}$}_{t}^k}\}_{k=1}^K)
\end{equation}
where $\tilde{\mathcal{M}}_i^k$ refers to the refined mask for the $k^{\text{th}}$ object at frame $i$. 

In overview, a variant of a transformer decoder is used, 
as illustrated in Figure~\ref{fig:rgb_corrector}. A query vector is provided for each object, initialised from the exemplar masks for that object, and subsequently the object query vectors are jointly processed, and attend to the target and exemplar mask frames
to establish temporal correlations in the visual stream. Again, DINO features are used to enable semantic grouping and temporal correspondence. In the following, we give more details for each element of this architecture.

{\noindent \bf Query Initialisation. } 
Foreground object-wise query vectors~($q_{\text{obj}} \in \mathbb{R}^{K \times D}$) are initialised from 
the visual features for the set of exemplar mask frames, 
by weighted pooling the dense feature maps with the exemplar masks along both spatial and temporal dimensions. An additional set of $K_{\text{bg}}$ learnable embedding vectors~($q_{\text{bg}} \in \mathbb{R}^{K_{\text{bg}} \times D}$) models the background regions.
As a result, there are a total of $K+K_{\text{bg}}$ query vectors to represent the entire video scene. 
To account for the temporal dimension in videos, 
we broadcast these query vectors to $T$ target frames, 
leading to a total of ($K + K_{\text{bg}}) \times T$ initialised query vectors, {\em i.e.,}~$q \in \mathbb{R}^{(K + K_{\text{bg}}) \times T \times D}$.

{\noindent \bf Transformer Module. } 
Once initialised, the queries $q$ undergo iterative refinements through a series of transformer-like layers, with two differences to the standard transformer decoder: (i) we insert an additional multi-head cross-attention layer;
(ii) we apply a Slot-Attention-like mechanism~\cite{locatello2020object} within each cross-attention module, encouraging competition between foreground objects and background regions.

In the first cross-attention layer, the query vectors attend to the visual features of all adjacent \textbf{target} frames, whereas in the second cross-attention layer, 
the key and value entries are generated by combining \textbf{exemplar} masks with the visual features of corresponding frames. These layers enable the incorporation of short-term local information and global exemplar information, respectively.

{\noindent \bf Refined Mask Outputs. }
Following a similar idea to~\cite{cheng2021maskformer, Xie22}, 
we extract the cross-attention maps between query vectors and visual features of target frames from the last layer of the transformer module.
Notably, we only retain the attention maps that are associated with $K$ foreground object queries.
These filtered attention maps are then upsampled by a CNN decoder to yield refined masks {\small $\{\tilde{\mathcal{M}}_{1}^k, \dots, \tilde{\mathcal{M}}_{T}^k\}_{k=1}^K$}.
During the upsampling process, 
information of flow-predicted masks and object appearances in target frames are passed through a lightweight CNN encoder, acting as skip layers for fine-grained feature compensation.

{\noindent \bf Mask Prediction Loss.} At training time, we apply both binary cross-entropy loss and boundary loss (adopted from~\cite{Xie22}) on the resultant refined masks
\begin{equation}
    L_{\text{mask}} = \frac{1}{KT}\sum_{i=1}^{T}\sum_{k=1}^K 
    \Big( \Big.  \lambda_{\text{bce}}L_{\text{bce}}(\tilde{\mathcal{M}}_{i}^k, {\mathcal{M}}_{i}^k) 
    \label{loss:mask}
    + \lambda_{\text{bound}}L_{\text{bound}}(\tilde{\mathcal{M}}_{i}^k, {\mathcal{M}}_{i}^k) \Big. \Big)
\end{equation}
with $\lambda_{\text{bce}}, \lambda_{\text{bound}}$ denoting weight factors. The groundtruth segmentation masks {\small $\{{\mathcal{M}}_{i}^k\}$} are provided by synthetic video sequences, as will be detailed in Sec.~\ref{sec:train}.

\vspace{0.15cm}
{\noindent \bf Feature Reconstruction Loss. } 
In addition to direct supervision, we incorporate feature reconstruction as an auxiliary task. 
Specifically, the output query vectors from the transformer module are spatially broadcast and projected by MLP layers to reconstruct the DINO feature maps $d$ of input frames. This reconstruction process is inspired from~\cite{Yang21a, seitzer2023bridging}, where the feature predictions by each query {\small $\hat{d}^k$} are linearly combined, weighted by the predicted alpha channels {\small $\hat{\alpha}^k$}, {\em i.e.,~}{\small $\hat{d} = \sum_{k} \hat{\alpha}^k \hat{d}^k$}.

The overall loss for this feature reconstruction process is averaged over all target frames:
\begin{equation}
    L_{\text{feat-recon}} = \frac{1}{T}\sum_{i=1}^T \Big( \Big. \lambda_{\text{recon}}\text{MSE}(\hat{d}_i, d_i)
    + \frac{1}{K}\sum_{k=1}^K \lambda_{\text{alpha}} L_{\text{bce}}(\hat{\alpha}_i^k, \alpha_i^k) \Big. \Big)
    \label{loss:featrecon}
\end{equation}
where the first term (weighted by $\lambda_{\text{recon}}$) denotes an unsupervised feature reconstruction loss, and the second term (weighted by $\lambda_{\text{alpha}}$) corresponds to a supervised alpha channel prediction loss. This supervision signal is only applied to the alpha masks generated from foreground object queries, with the groundtruth {\small $\{\alpha_i^k\}$} obtained by downsampling the groundtruth segmentation masks {\small $\{\mathcal{M}_i^k\}$}.

\subsection{Sim2Real Training Procedure} 
\label{sec:train}

Training proceeds in two steps: first, the selector and corrector are pre-trained with synthetic videos; and second, a self-supervised adaptation process is used to bridge the domain gap to real-world videos. Both steps are described below.
Notably, this entire training process does not require any manual annotations.

\subsubsection{Pre-training on Synthetic Videos.}
We adopt the simple simulation engine proposed by~\cite{Xie22} to generate synthetic training video sequences with RGB frames, optical flows and groundtruth annotations.

{\noindent \textit{Synthetic Training of Mask Selector. }} 
Given one synthetic video sequence, we feed its optical flows into the OCLR-flow multi-object motion segmentation model, to yield flow-predicted masks as the selector inputs. The groundtruth error maps for training (as introduced in Eq.~\ref{loss:errormap}) can be obtained by comparing the flow-predicted masks with groundtruth segmentation maps. 

{\noindent \textit{Synthetic Training of Mask Corrector. }}
By combining loss functions introduced in the refined mask prediction (Eq.~\ref{loss:mask}) and feature reconstruction (Eq.~\ref{loss:featrecon}) processes, the overall loss for training the mask corrector is
\begin{equation}
    \label{loss:corrector}
    L_{\text{corrector}} = \lambda_{\text{mask}} L_{\text{mask}} +  \lambda_{\text{feat-recon}} L_{\text{feat-recon}}
\end{equation}
where $\lambda_{\text{mask}}$ and $\lambda_{\text{feat-recon}}$ are corresponding weight factors, and the supervision is provided by the groundtruth annotations in the synthetic sequences.

Both the selector and corrector models, trained on synthetic data, are able to generalise to real-world videos, as will be demonstrated in our ablation analysis.

\vspace{-0.2cm}
\subsubsection{Self-Supervised Adaptation to Real-World Videos.}
\label{subsec:ttt}

The Sim2Real domain gap is mitigated by self-supervised training on target real-world datasets.
Specifically, for each real-world video dataset, we first employ our synthetically-trained selection-correction model to generate segmentation masks starting from the OCLR-flow proposals. The resultant masks are then treated as pseudo-labeling to fine-tune the mask corrector and the OCLR-flow segmentation model. We provide more details on each fine-tuning process below, and verify the effectiveness of each stage via ablation analysis in in~\Cref{supsubsec:ablation-ssa}.

\vspace{0.15cm}
{\noindent \textit{Adapting the Mask Correction Model.} }
The overall loss for fine-tuning remains the same as for synthetic training (Eq.~\ref{loss:corrector}). However, for adaptation, the refined masks generated by the synthetically-trained corrector {({\em i.e.,~}{\small $\{\tilde{\mathcal{M}}_{i}^k\}$})} are treated as pseudo-GT, and become the target masks in the loss.
{Note that, we only apply pseudo-labeling at the frames where exemplar masks present due to their high quality. }
Apart from the self-supervised training at the dataset level, we also employ a test-time per-sequence adaptation to further boost the performance. A detailed description is given in~\Cref{supsec:implementation}.

\vspace{0.15cm}
{\noindent \textit{Adapting the Flow Segmentation Model.} }
Similarly, the OCLR-flow model (which is also originally trained only on synthetic data) is adapted to real-world videos by fine-tuning with the pseudo-GT provided by the refined segmentation masks {\small $\{\tilde{\mathcal{M}}_{i}^k\}$}.
This adapts the model to complex real-world flow inputs, consequently reducing Sim2Real domain gap.

The updated flow-predicted masks can then form the basis for a further appearance-based selector-corrector refinement. In principle, it would be possible to repeat this refinement-adaptation process. However, in practice, we find that one round of adaptation is sufficient and there is little gain in iterating further.

\section{Experiments}
\label{sec:experiment}
\subsection{Datasets and Metrics}
\noindent {\bf Training Dataset. } To pre-train our models, we use a synthetic dataset generated by the simple pipeline introduced in~\cite{Xie22}, where motions of multiple object sprites ({\em i.e.,~}shapes with random textures) are simulated on a common moving background across time. Each sequence contains $1$ to $3$ moving sprites, with information such as RGB frames, optical flows, groundtruth modal and amodal annotations provided. 

\vspace{0.1cm}
\par \noindent {\bf Multi-Object Benchmark. }
We adopt the DAVIS2017-motion dataset (short form ``DAVIS17-m'') that is proposed in~\cite{Xie22} with jointly moving objects re-annotated. We further formulate a YouTubeVOS2018-motion dataset (short form ``YTVOS18-m''), which is a subset of YTVOS18~\cite{vos2018}, containing $120$ sequences with one or more dominant moving objects. Visualisations of multi-object sequences can be found in Figure~\ref{fig:multivos}. Additional information regarding this new benchmark is provided in~\Cref{supsubsec:datasets_ytvos}.

\vspace{0.1cm}
\noindent {\bf Single-Object Benchmark. }
We report performance on popular single-object video segmentation benchmarks including DAVIS2016~\cite{Perazzi16}, SegTrackv2~\cite{FliICCV2013} and FBMS-59~\cite{OB14b}. Though some sequences in SegTrackv2 and FBMS-59 consist of multiple objects, we follow the common practice~\cite{Jain17,yang_loquercio_2019} and group foreground objects as a whole for later evaluation.

\vspace{0.1cm}
\noindent {\bf Evaluation Metrics. }
We benchmark our single object video segmentation performance using region similarity ($\mathcal{J}$). For multi-object video segmentation, we additionally report the contour similarity ($\mathcal{F}$)~\cite{1273918}.

\subsection{Implementation Details}
For pre-processing, we apply RAFT~\cite{Teed20} to estimate optical flow with frame gaps $\pm 1$ (except for FBMS and YTVOS18-m with frame gaps $\pm 3$ owing to relatively slow video motions). All optical flows and RGB frame inputs are resized to $128 \times 224$. 
The motion segmentation model proposes $K = 3$ objects for each sequence,  
from which $t = 10$ exemplar masks are selected for each object. 
Also, we use self-supervised pre-trained DINO transformers~\cite{caron2021emerging}, ViT-S/16 for mask selector, ViT-S/8 for corrector, resulting in DINO feature resolutions $8 \times 14$ and $16 \times 28$, respectively. 
As a final post-processing step, we apply Conditional Random Field (CRF) segmentation to further refine the output masks.
We refer the reader to~\Cref{supsec:impl} for more details regarding the model architecture, training settings, and inference time analysis.

\subsection{Multiple Object Video Segmentation}
\begin{table}[!t]
\centering
\setlength\tabcolsep{4pt}
\resizebox{0.98\textwidth}{!}{
  \begin{tabular}{lccccccccccc}  
    \toprule
    {} & \multicolumn{5}{c}{Model settings} & \multicolumn{3}{c}{YTVOS18-m} & \multicolumn{3}{c}{DAVIS17-m}  \\
    \cmidrule(r){2-6}
    \cmidrule(r){7-9}
    \cmidrule(r){10-12}
    Model    & VOS & H.A. & RGB &  Flow & Input Res.  & {$\mathcal{J}\&\mathcal{F}$  $\uparrow$} & {$\mathcal{J}$ $\uparrow$} & {$\mathcal{F}$ $\uparrow$} & {$\mathcal{J}\&\mathcal{F}$  $\uparrow$} & {$\mathcal{J}$ $\uparrow$} & {$\mathcal{F}$  $\uparrow$}\\
    \midrule
    $^{\dag}$MG~\cite{Yang21a} & un-sup. & \xmark  & \xmark & $\checkmark$ & $128 \times 224$ & $33.3$ & $37.0$ & $29.6$ &  $35.8$ & $38.4$  &  $33.2$    \\
    OCLR-flow~\cite{Xie22} &  un-sup. & \xmark & \xmark  &$\checkmark$ & $128 \times 224$ & $45.3$ & $46.5$ & $44.1$ &  $55.1$ &  $54.5$  &  $55.7$     \\
    {Safadoust et al.~\cite{Safadoust23}}  &  un-sup. & \xmark &  $\checkmark$ &  \xmark & $128 \times 224$ & $-$ & $-$ & $-$ &  $59.2$ &  $59.3$  &  $59.2$     \\
    OCLR-TTA~\cite{Xie22} & un-sup. & \xmark & $\checkmark$  & $\checkmark$ & $480 \times 854$ & $50.6$ & $52.7$ & $48.6$  &  $64.4$ &  $65.2$  &  $63.6$   \\
    {$^{\dag}$VideoCutLER~\cite{wang2023videocutler}} &  un-sup. & \xmark &  $\checkmark$ & \xmark & $480 \times 854$ & $57.0$ & $59.0$ & $55.1$ &  $57.3$ &  $57.4$  &  $57.2$     \\
    \textbf{Ours} & un-sup.  & \xmark & $\checkmark$  & $\checkmark$ & $128 \times 224$ & $\textbf{65.2}$ & $\textbf{67.1}$ & $\textbf{63.2}$ &  $\textbf{66.2}$ &  $\textbf{67.0}$  &  $\textbf{65.4}$   \\
    \midrule
    {OCLR-flow~\cite{Xie22} + SAM~\cite{kirillov2023segany}} &  un-sup. & $\checkmark$ & $\checkmark$  & $\checkmark$ & $576 \times 1024$ & $58.5$ & $57.0$ & $60.0$ &  $64.2$ &  $62.0$  &  $66.4$     \\
    {\textbf{Ours} + SAM~\cite{kirillov2023segany}} &  un-sup. & $\checkmark$ & $\checkmark$  & $\checkmark$ & $576 \times 1024$ & $\textbf{70.6}$ & $\textbf{71.1}$ & $\textbf{70.2}$ &  $\textbf{71.5}$ &  $\textbf{70.9}$  &  $\textbf{72.1}$     \\
    \midrule
    $^{\dag}$CorrFlow~\cite{Lai19} & semi-sup. & \xmark & $\checkmark$  &  \xmark & $480 \times 854$ & $55.2$ & $60.0$ & $50.4$  &  $54.0$ &  $54.2$  &  $53.7$   \\
    $^{\dag}$UVC~\cite{nips19_joint_task} & semi-sup. & \xmark & $\checkmark$  &  \xmark & $480 \times 854$ & $66.9$ & $70.4$ & $63.4$  &  $65.5$ &  $66.2$  &  $64.7$  \\
    $^{\dag}$CRW~\cite{jabri2020walk} & semi-sup. & \xmark & $\checkmark$  & \xmark & $480 \times 854$ & $72.9$ & $75.5$ & $70.4$ & $73.4$ &  $72.9$  &  $74.1$ \\
    $^{\dag}$DINO~\cite{caron2021emerging} & semi-sup. & \xmark & $\checkmark$  &  \xmark & $480 \times 854$ & $\textbf{75.4}$ & $\textbf{76.6}$ & $\textbf{74.3}$  & $\textbf{78.7}$ &  $\textbf{77.7}$  &  $\textbf{79.6}$  \\
    \bottomrule
  \end{tabular}}
  \caption{\textbf{Quantitative comparison on multi-object video segmentation benchmarks.} ``H.A.'' indicates that human annotations are involved during training. {$\dag$ denotes that we reproduce the results on the benchmark datasets using the official code released by the original authors.} In the semi-supervised setting, the first-frame ground-truth annotation is provided.}
  \label{tab:multivos}
\end{table}

\begin{figure}[!tbph]
    \centering
    \includegraphics[width=0.95\textwidth]{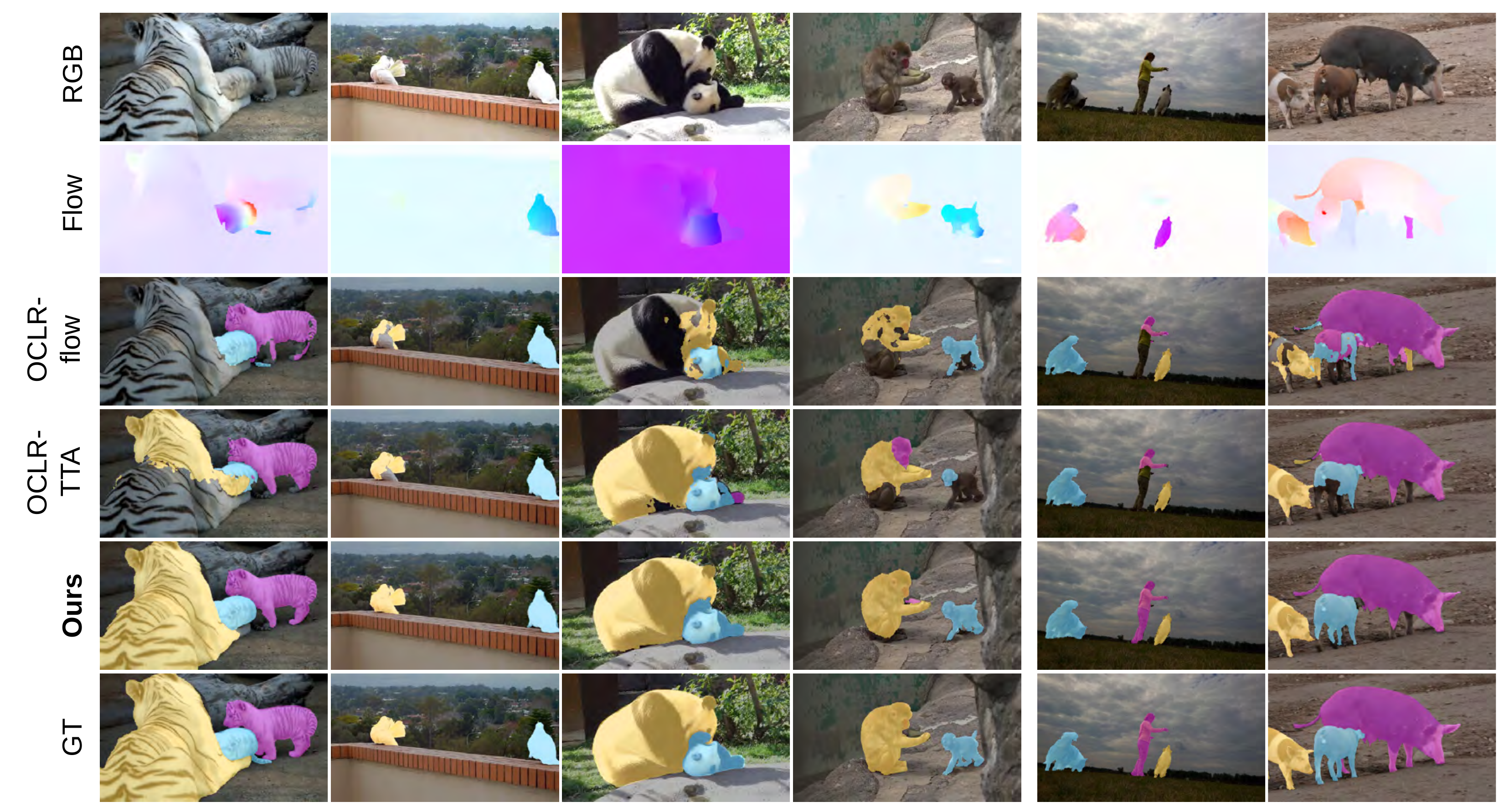}
    \caption{\textbf{Qualitative comparison on multi-object video segmentation benchmarks}, including YTVOS18-m (left) and DAVIS17-m (right).}
    \label{fig:multivos}
    \vspace{-0.1cm}
\end{figure}

\noindent \textbf{Comparison to Unsupervised Models.}
Table~\ref{tab:multivos} compares the performance of 
our motion segmentation model against several baselines on unsupervised video object segmentation benchmarks.
Interestingly, most prior baseline methods exhibit degraded performance on the YTVOS18-m dataset compared to the DAVIS17-m dataset. This is understandable, as the former dataset involves more complicated motion patterns, posing difficulties for motion-based segmentation. However, our method effectively handles this problem and demonstrates superior performance on both benchmark datasets.

Our findings are further supported by comprehensive ablation studies provided in~\Cref{supsec:ablation}. These studies verify our hyperparameter choices (\emph{e.g.}, the number of exemplar masks) and architectural designs (\emph{e.g.}, the choice of self-supervised features) during the synthetic training phase. Moreover, we detail how various self-supervised adaptation stages contribute to the overall performance.

\vspace{0.1cm}
\noindent \textbf{Comparison to the Segment Anything Model (SAM)~\cite{kirillov2023segany}.}
Recent studies highlight SAM as a robust general segmentor with diverse prompt options. To assess its efficacy for refining flow-based segmentation, we prompt SAM with masks predicted by the OCLR-flow model in a per-frame manner. As shown in Table~\ref{tab:multivos}, the resultant refinements (OCLR-flow + SAM) are not as effective as the refinements provided by our selection-correction model (Ours), owing to the limited utilisation of temporal consistency in the former approach. On the other hand, prompting SAM with masks predicted by our model does lead to improvements (Ours + SAM). This demonstrates that the SAM-based refinement is complementary to our method built on the persistence of appearance streams. Visualisations of SAM-based refinements are given in~\Cref{supsec:qualitative}.

\vspace{0.1cm}
\noindent \textbf{Comparison to Semi-Supervised Models.} Table~\ref{tab:multivos} also includes a list of existing semi-supervised methods that were re-run on motion segmentation benchmarks for reference. Interestingly, the issue of stationary frames in YTVOS18-m actually benefits semi-supervised tasks, as objects that move infrequently are easier to track using first-frame ground truth. Despite this advantage, our unsupervised-VOS method still achieves competitive results compared to some semi-supervised-VOS baselines.

\vspace{0.1cm}
\noindent \textbf{Qualitative Results.} Figure~\ref{fig:multivos} presents visualisations of multi-object sequences with complex flow signals. It is evident that the method relying solely on flow inputs ({\em i.e.,~}OCLR-flow) sometimes fails to extract object shapes from static flow fields and articulated motion signals. OCLR-TTA addresses this issue by using RGB-based dense mask propagations, but still struggles with completing the shape of objects, such as the large white tiger in the first column. In contrast, our method accurately recovers object segmentation even under extreme motion scenarios, thanks to its object-centric representation and reconstruction-based scene clustering.

\begin{table}[t]
  \centering
  \setlength\tabcolsep{11pt}
  \resizebox{0.98\textwidth}{!}{
  \begin{tabular}{lcccccccc}  
    \toprule
    {} & \multicolumn{4}{c}{Model settings} & \multirow{2}[2]{*}{\shortstack{DAVIS16 \\ $\mathcal{J} \uparrow$}} & \multirow{2}[2]{*}{\shortstack{STv2 \\ $\mathcal{J} \uparrow$}}  & \multirow{2}[2]{*}{\shortstack{FBMS59 \\ $\mathcal{J} \uparrow$}}  \\
    \cmidrule(r){2-5}
    Model    &  H.A.  & RGB &  Flow  &  Input Res. &  &  &\\
    \midrule
    MG~\cite{Yang21a} & \xmark & \xmark  &$\checkmark$ & $128 \times 224$ &  $68.3$ &  $58.6$  &  $53.1$ \\
    $^{\ddagger}$OCLR-flow~\cite{Xie22}& \xmark  & \xmark  &$\checkmark$ & $128 \times 224$ &  $72.1$ &  $67.6$  &  $70.0$ \\
    DS~\cite{ye2022sprites} & \xmark  &$\checkmark$  &$\checkmark$ & $240 \times 426$ &  $79.1$ &  $72.1$ & $71.8$ \\
    DystaB~\cite{Yang_2021_CVPR} & \xmark  &$\checkmark$  &$\checkmark$ & $192 \times 384$ &  $80.0$ &  $74.2$ & $73.2$ \\
    GWM~\cite{Choudhury22} & \xmark  &$\checkmark$  & \xmark &  $128 \times 224$  &  $80.7$ &  $78.9$ & $78.4$ \\
    $^{\ddagger}$OCLR-TTA~\cite{Xie22} & \xmark  &$\checkmark$  &$\checkmark$ &  $480 \times 854$ &  $80.9$ &  $72.3$ & $72.7$ \\
    {LOCATE~\cite{LOCATE}} & \xmark  &$\checkmark$  & \xmark &  $480 \times 848$ &  $80.9$ &  $\textbf{79.9}$ & $68.8$ \\
    $^{\ddagger}$\textbf{Ours} & \xmark  &$\checkmark$  &$\checkmark$ &  $128 \times 224$ &  $\textbf{81.1}$ &  $76.6$ & $\textbf{81.9}$ \\
    \midrule
    {OCLR-flow~\cite{Xie22} + SAM~\cite{kirillov2023segany}} & $\checkmark$ & $\checkmark$  & $\checkmark$ & $576 \times 1024$ & $80.6$ & $71.5$ & $79.2$    \\
    MATNet~\cite{zhou20} & $\checkmark$  & $\checkmark$  &$\checkmark$ & $473 \times 473$ &  $82.4$  &  $50.4$   &  $76.1$    \\
    DystaB~\cite{Yang_2021_CVPR} & $\checkmark$  &$\checkmark$  &$\checkmark$ & $192 \times 384$ &  $82.8$ &  $74.2$ & $75.8$ \\
    {\textbf{Ours} + SAM~\cite{kirillov2023segany}} & $\checkmark$ & $\checkmark$  & $\checkmark$ & $576 \times 1024$ & $86.6$ & $\textbf{81.3}$  & $\textbf{85.7}$      \\
    DPA~\cite{cho2023dual} & $\checkmark$  & $\checkmark$  & $\checkmark$ & $512 \times 512$ &  $\textbf{87.1}$  &  $-$   &  $81.0$ \\
    \bottomrule
  \end{tabular}}
  \caption{\textbf{Quantitative comparison on single object video segmentation benchmarks.} ``H.A.'' indicates that human annotations are involved during training. $\ddagger$ corresponds to methods that rely on human-label-free supervision on synthetic data.}
  \label{tab:singlevos}
  \vspace{-0.3cm}
\end{table}

\subsection{Single Object Video Segmentation}

We also summarise our performance on various single-object segmentation benchmarks.
These comparisons with other self-supervised approaches are considered reasonable since both setups rely on unsupervised learning schemes on targeted data distributions, and no manual annotations are used during the training process.
While our method is not specialised for the task of single object segmentation, as demonstrated in Table~\ref{tab:singlevos}, it still achieves the top performance on two out of three datasets among methods that do not use human annotations. It does less well on the SegTrack dataset since the low-quality sequences (that have watermarks and video compression artefacts) result in noisy flow estimates that compromise our performance.

\begin{figure}[!t]
    \centering
    \includegraphics[width=0.98\textwidth]{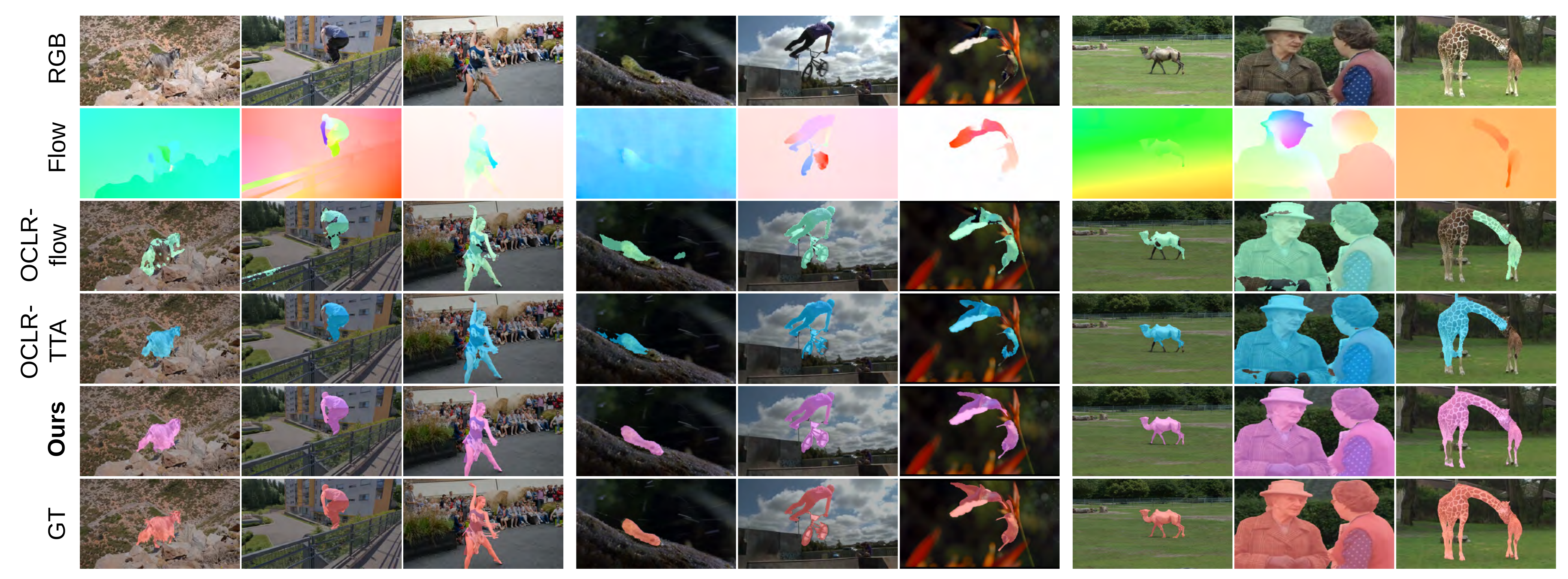}
    \caption{\textbf{Qualitative comparison on single object video segmentation benchmarks}, including DAVIS16 (left), 
    SegTrackv2 (middle), and FBMS-59 (right).}
    \label{fig:singlevos}
\end{figure}

Furthermore, applying per-frame SAM to refine mask predictions, gives noticeable improvements, and leads to the top or matching performance among supervised methods.
Figure~\ref{fig:singlevos} provides corresponding visualisations on single-object benchmarks. Similar to the multi-object case, our method exhibits outstanding competence in handling challenging motion situations.

\section{Discussion}
\label{sec:discussion}
In this paper, we introduce an appearance-based refinement method to address inaccuracies in flow-predicted masks. Our approach involves synthetic pre-training, followed by self-supervised adaptation to real-world videos, without relying on any human annotations. By formulating a new motion segmentation benchmark, we have demonstrated the efficacy of our method in handling complicated motion patterns such as moving objects that are motionless for a time, environment motions, multi-object interactions, {\em etc}. 

Instead of pursuing an end-to-end design, we have explored the potential of employing multiple stages: 
flow-based mask proposals, appearance-based mask selection, and appearance-based mask correction, with each stage utilising a lightweight interpretable model. More importantly, this approach allows us to efficiently harness the advantages of both flow and appearance modalities: the object discovery and accurate boundary information from optical flow; and the temporal consistency observed in appearance streams.

Nevertheless, some limitations have also been revealed under extremely complex motion scenarios: \textit{First}, when a significant majority of problematic flow-predicted masks are present ({\em e.g.,~}due to a large proportion of stationary frames in the sequence), selecting meaningful 
{exemplar masks} becomes challenging, consequently affecting the correction process. This problem could be potentially mitigated by incorporating additional appearance priors into the flow-based proposal stage. \textit{Second}, the adopted frozen DINO ViT mainly encodes semantic-level information, which could result in difficulties in discerning entangled objects with similar appearances at the instance level.  This highlights a possible research direction, aiming to develop instance-level appearance representations by leveraging flow-predicted information in complex multi-object scenes.
\section*{Acknowledgements}
This research is supported by the UK EPSRC Programme Grant Visual AI (EP/T028572/1), a Royal Society Research Professorship RP$\backslash$R1$\backslash$191132, and a Clarendon Scholarship.

%
%
\bibliographystyle{splncs04}
\bibliography{main,vgg_local}

\newpage
\appendix
\begin{center}
    {\large \textbf{Appendix}}
\end{center}
\noindent This appendix contains the following parts:
\begin{itemize}
    \item \textbf{\cref{supsec:impl}: Architecture and Implementation Details,} where we provide more information regarding architectural designs and implementation details.

    \item \textbf{\cref{supsec:datasets}: Datasets,} where we provide a detailed description of all datasets adopted for training and evaluation.
    
    \item \textbf{\cref{supsec:ablation}: Ablation Studies,} where we conduct a comprehensive ablation analysis of the key designs and components of the model.

    \item \textbf{\cref{supsec:quantitative}: Additional Quantitative Results,} where complete quantitative comparisons are provided between our method and prior works.

    \item \textbf{\cref{supsec:qualitative}: Additional Qualitative Results} are also provided on both final predictions and intermediate results.
    
    \item \textbf{\cref{appendix:ethic}: Potential Societal Impacts} of our work are discussed.
\end{itemize}
\vspace{0.cm}
\section{Architecture and Implementation Details}
\label{supsec:impl}
\subsection{Appearance-Based Mask Selector}
\begin{figure}[hbtp]
    \hspace{0.25cm}
    \includegraphics[width=0.7\textwidth]{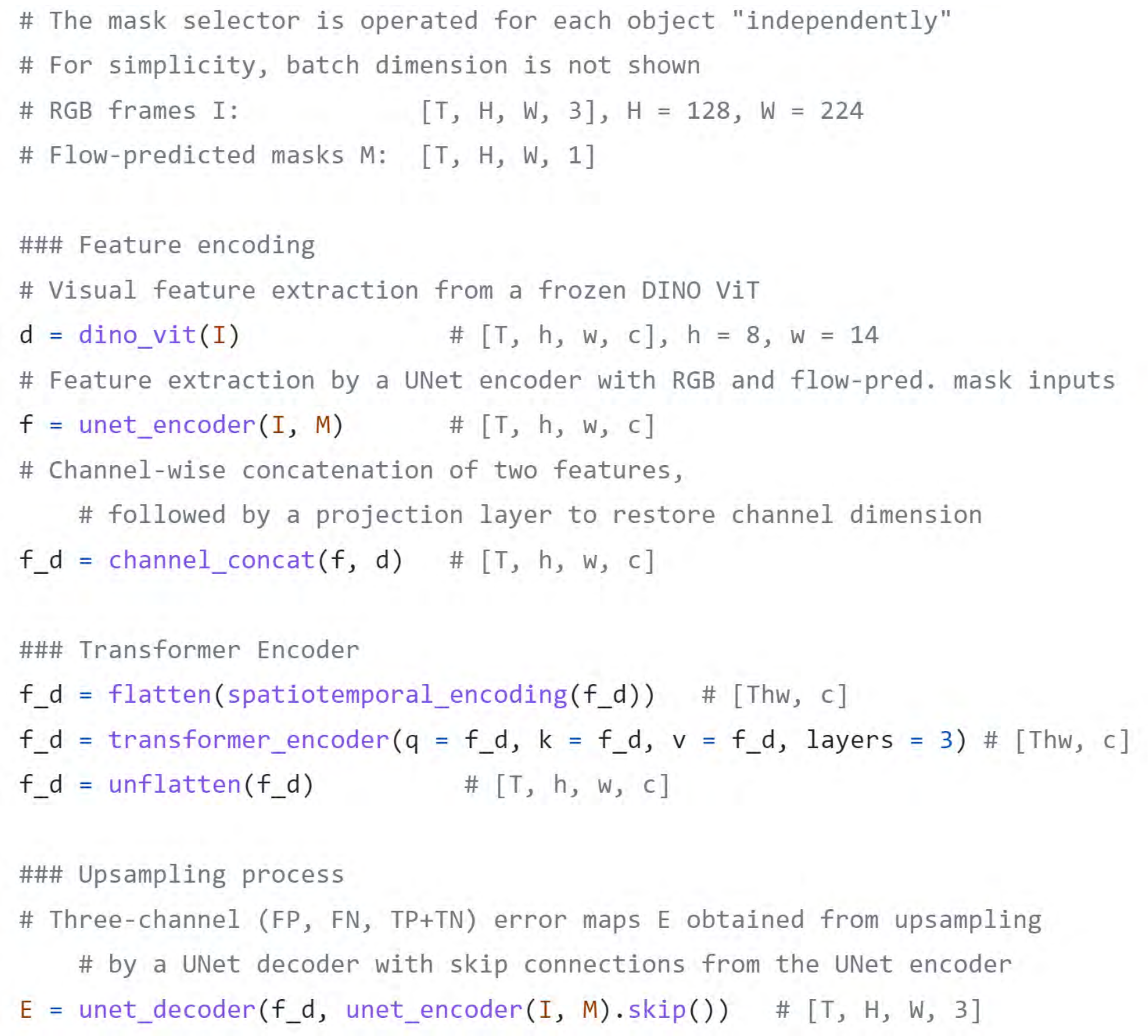}
    \caption{\textbf{Pseudo-code for the mask selector.}}
    \label{supfig:code_selector}
    \vspace{-0.5cm}
\end{figure}

Figure~\ref{supfig:code_selector} presents the pseudo-code for the mask selector that takes in both RGB frames and flow-predicted masks, and outputs framewise error maps (with three channels corresponding to FP, FN, and TP+TN regions) for each object \textbf{independently}.  

\vspace{3pt}
\par{\noindent \textbf{Exemplar Mask Selection. }} 
Given an object $k$, for each frame $i$,  we compute the sum of erroneous regions ({\em i.e.,~}the total number of pixels in FP and FN regions) in the error map, denoted as {\small $A_{\text{error}, i}^k \in \mathbb{N}$}. 
We then take the ratio between {\small $A_{\text{error}, i}^k$} and the area of the flow-predicted mask {\small $A_{i}^k$}, which leads to a relative erroneous area {\small $a_{\text{error}, i}^k = A_{\text{error}, i}^k / A_{i}^k$} for each frame.

These relative erroneous areas are treated as indications of the mask quality. Intuitively, for a fixed size of error regions {\small $A_{\text{error}, i}^k$}, the flow-predicted mask with a larger area {\small $A_{i}^k$} would be considered as more accurate, which can be indicated by a smaller relative erroneous area {\small $a_{\text{error}, i}^k$}. 

We compare the relative erroneous areas across frames to obtain a ranking of the mask quality, 
and the top-$t$ exemplar masks are then selected for each object. 

\subsection{Appearance-Based Mask Corrector}
\begin{figure*}[hbtp]
    \hspace{0.cm}
    \includegraphics[width=0.99\textwidth]{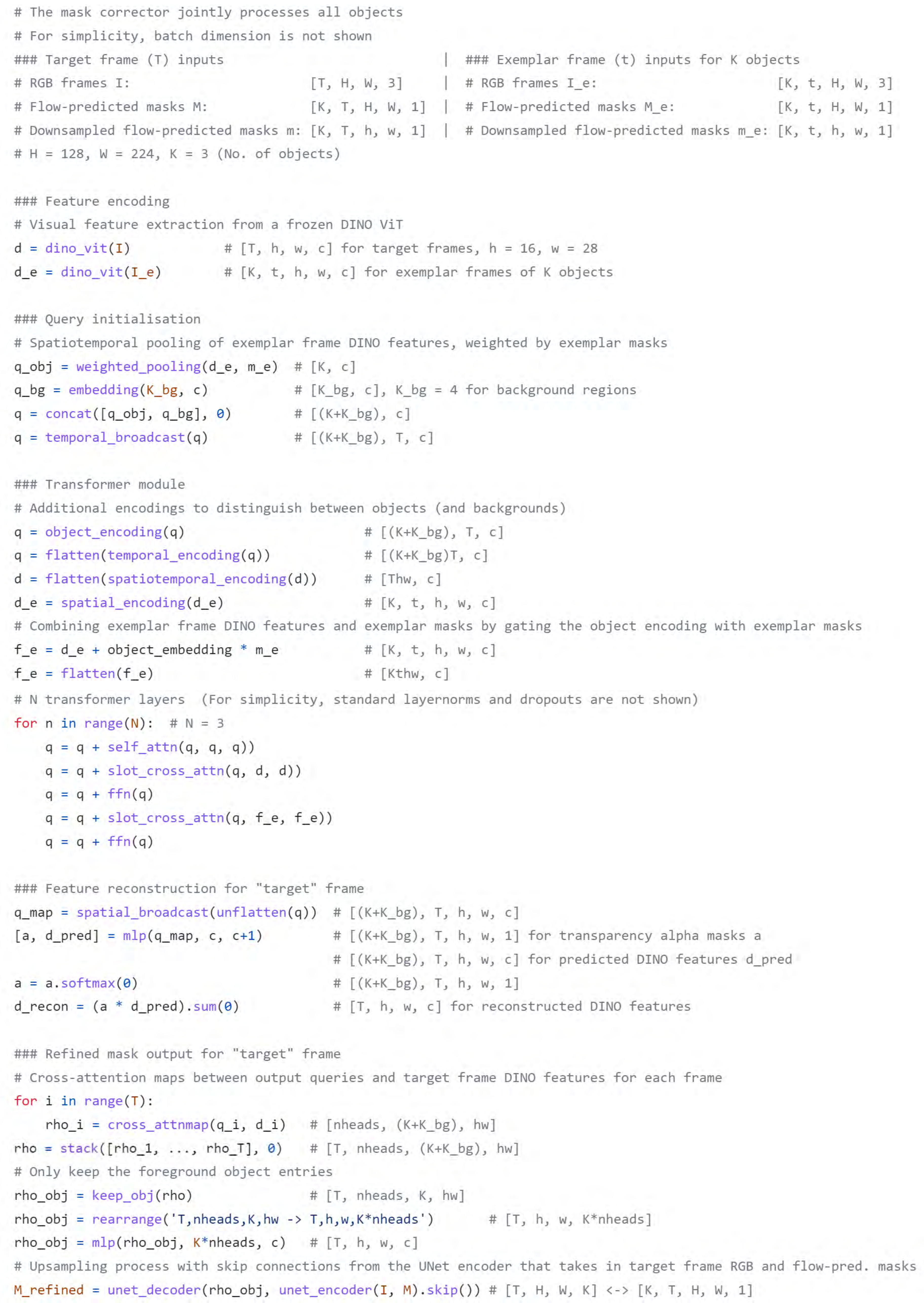}
    \caption{\textbf{Pseudo-code for the mask corrector.}}
    \label{supfig:code_corrector}
\end{figure*}

We also provide the pseudo-code for the mask corrector in Figure~\ref{supfig:code_corrector}, where we outline the major stages including DINO feature extraction, query initialisation, transformer module, feature reconstruction, and refined mask output.

\subsection{Implementation Details}
\label{supsec:implementation}
\par{\noindent \textbf{Input Settings.}} During pre-processing, we apply RAFT~\cite{Teed20} to estimate optical flow with frame gaps $\pm 1$ (except for FBMS and YTVOS18-m with frame gaps $\pm 3$ due to relatively slow video motions). All optical flows and RGB frame inputs are resized to $128 \times 224$. We adopt self-supervised pre-trained DINO transformers~\cite{caron2021emerging},
ViT-S/16 for mask selector, ViT-S/8 for corrector, resulting in DINO feature resolutions $8 \times 14$ and $16 \times 28$, respectively.
\vspace{3pt}
\par{\noindent \textbf{Model Settings. }} For flow-based segmentation, we follow the original OCLR-flow model settings and process $30$ optical flow frames as a batch, with $K = 3$ objects predicted for each frame. 
The mask selector is designed to handle flow-predicted masks for up to $100$ frames, 
from which $t = 10$ exemplar masks are selected for each object.
After that, we divide the entire sequence into non-overlapping temporal windows, 
each consisting of $T = 7$ frames, to be fed into the subsequent mask corrector as target frames. 
Within the mask corrector, we use $K_{\text{bg}} = 4$ learnable embeddings to represent the background components, leading to a total of $7$ initialised queries ($3$ object proposals + $4$ background regions) explaining the whole video scene. 
\vspace{3pt}
\par{\noindent \textbf{Trainable Parameters.}}
The proposed appearance refinement models are light-weight, totaling $19.5 M$ trainable parameters ($3.3 M$ parameters for the mask selector and $16.2 M$ parameters for the mask corrector). During the self-supervised adaptation, we fine-tune the OCLR-flow model, which comprises $20.7 M$ trainable parameters.
\vspace{3pt}
\par{\noindent \textbf{Synthetic Pre-training Settings.} }
The selector and corrector are independently trained from scratch with the Adam optimizer~\cite{Adam}. The learning rate is linearly warmed up to $5 \times 10^{-5}$ during the initial $40$k iterations and subsequently decayed by $50\%$ every $80$k iteration.
Both models are trained on a single NVIDIA Tesla V100 GPU, with each requiring roughly $4$ days to reach full convergence.

\vspace{3pt}
\par{\noindent \textbf{Self-Supervised Adaptation Settings.}}
We apply self-supervised training for both the mask corrector and the flow segmentation model on each target dataset, with both models converging in roughly a few hours. During the training, the learning rate is reduced to $2 \times 10^{-5}$. {As the pseudo-labels estimated by the refined masks could still be problematic ({\em e.g.,~}due to over-segmentation or missing parts), we remove the boundary loss component during the self-supervised training to prevent the model from focusing on erroneous boundaries.} 
{To adapt the mask corrector to real-world RGB frames, we zero out the flow-predicted mask inputs to the upsampling module, thereby forcing it to effectively utilise appearance inputs to recover the boundary details.} For the flow segmentation model, to avoid over-fitting to specific input flow patterns, we adopt different optical flow frame gaps for training and evaluation.

Furthermore, we fine-tune the mask corrector on each target video sequence for around $200$ iterations in a self-supervised manner, following the same settings as the dataset-level adaptation. This effectively serves as a \textit{per-sequence test-time adaptation} process.

{During inference, we apply the adapted models to real-world video sequences. The output masks are further refined by Conditional Random Field (CRF) segmentation as a final post-processing step.}

\vspace{3pt}
\par{\noindent \textbf{Inference Time Analysis.} }
The overall inference time of the model is summarised in Table~\ref{suptab:adap}, where the inference time taken for a forward pass (including flow-based proposal and appearance-based refinement) is $0.098s$ per frame.

The optional per-sequence test-time adaptation process described above provides a minor performance boost, typically less than $1\%$, as shown in Table~\ref{suptab:adap}. However, this extends the model running time to $2.030s$, though this time remains competitive with the inference time of the previous SOTA method ($2.202s$ for OCLR-TTA, excluding the CRF time).

\vspace{3pt}
\par{\noindent \textbf{Inference Settings for the Segment Anything Model (SAM)~\cite{kirillov2023segany}. } {SAM is applied as a per-frame appearance-based refinement method, wherein different object masks are processed as independent input prompts. Specifically, since SAM does not accept the mask prompt as the only input, we instead adopt a combination of the mask prompt and key point prompts for each instance mask.}
To generate the key point prompts for the corresponding mask, we sample approximately $10$ uniform grid points within the mask region for each instance. 
\clearpage
\clearpage
\section{Datasets}
\label{supsec:datasets}
\subsection{Synthetic Training Dataset}
We train our mask selector and corrector models on a simulated dataset introduced in~\cite{Xie22}. This training dataset consists of $4608$ sequences with $1$, $2$, and $3$ objects in equal proportions. Each sequence composes $30$ frames with information such as RGB frames, optical flows, groundtruth modal and amodal annotations provided. 

To generate foreground objects, random shapes ({\em e.g.,~}real object silhouettes and polygons) are textured by PASS images~\cite{asano21a} to form object sprites. Within each sequence, foreground sprites are overlaid on a common background following a particular order.

The motion of each object sprite is independently simulated by a random homography transformation and thin-plate spline mappings, whereas the background is either designed to move according to a homography or directly sampled from real background videos.

\begin{figure*}[hbtp]
    \hspace{0.1cm}
    \includegraphics[width=0.98\textwidth]{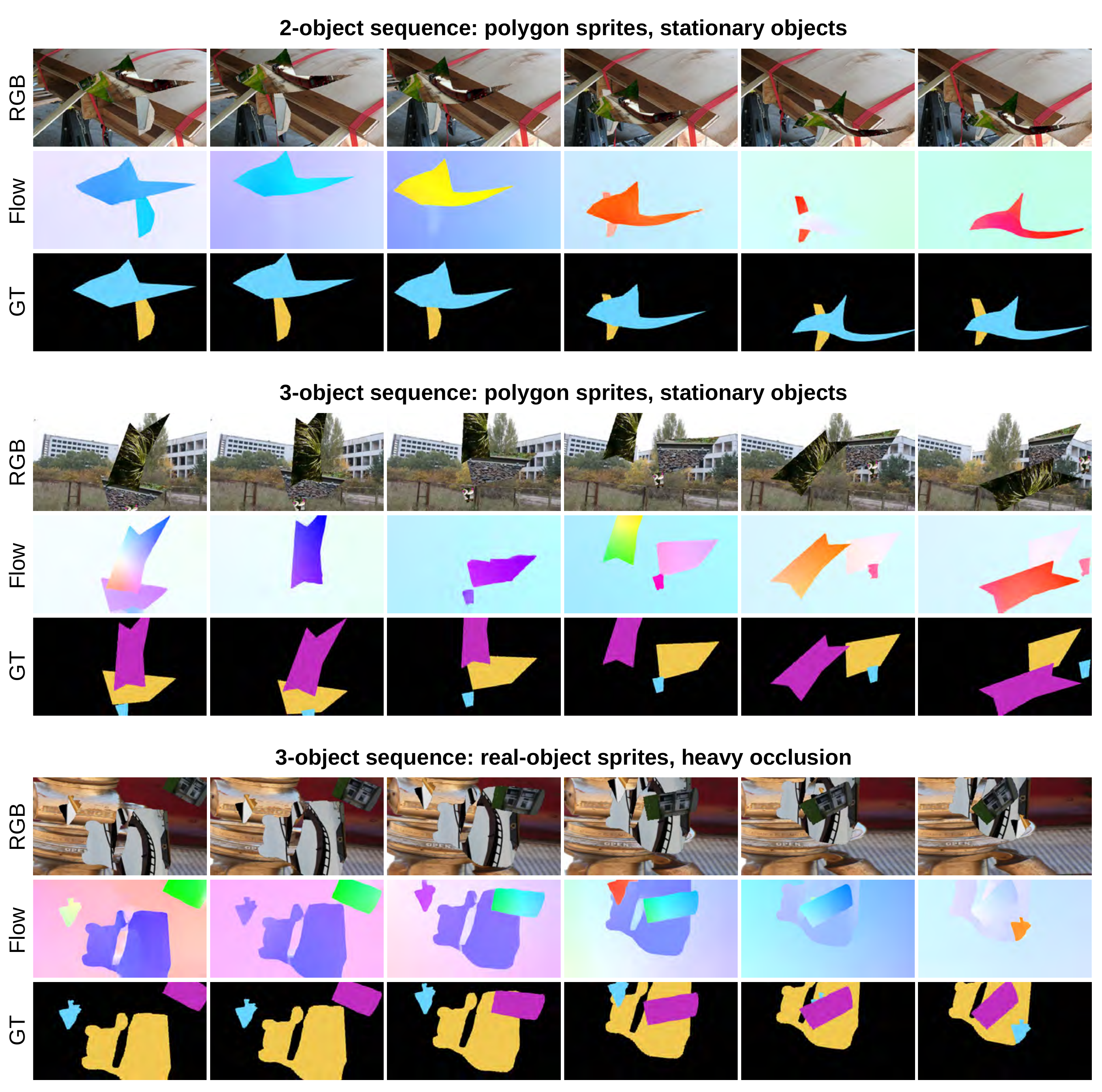}
    \caption{\textbf{Example synthetic sequences for training.} Every $5^{\text{th}}$ frame in the sequence is shown.}
    \hspace{-0.5cm}
    \label{supfig:syn}
\end{figure*}

Figure~\ref{supfig:syn} shows some example synthetic sequences where complex motion scenarios such as occlusions and stationary objects are simulated.

\subsection{Evaluation Datasets}
To evaluate our model performance, 
we adopt various benchmark datasets for both single and multiple object segmentation tasks. The details of these datasets are provided below.

\vspace{3pt}
{\noindent \bf DAVIS2016~\cite{Perazzi16}} is composed of $50$ video sequences at a $480$p resolution. Each sequence normally contains only one salient moving object in the scene. We report our model performance on its validation set with $20$ sequences.

\vspace{3pt}
{\noindent \bf SegTrackv2~\cite{FliICCV2013}} consists of $14$ sequences with a total of $947$ fully-annotated frames. The image quality and resolutions vary across different sequences. For some multi-object sequences, we follow the common practice~\cite{Jain17,yang_loquercio_2019} of single object video segmentation and combine all foreground annotations. 

\vspace{3pt}
{\noindent \bf FBMS-59~\cite{OB14b}} contains $59$ sequences with $720$ annotations sparsely provided at every $20$ frames. Similar to SegTrackv2, foreground objects in multi-object sequences are grouped together for single object segmentation evaluation.

\vspace{3pt}
{\noindent \bf DAVIS2017-motion~\cite{Xie22}} (short form ``DAVIS17-m'') is designed as a multi-object motion segmentation benchmark. It contains $30$ sequences that are curated from the original DAVIS2017~\cite{Ponttuset17} validation set. Specifically, annotations of jointly moving objects are combined, as objects in common motion are indistinguishable purely based on motion ({\em i.e.,~}flow) information.

\vspace{3pt}
{\noindent \bf YouTubeVOS2018~\cite{vos2018}} (short form ``YTVOS18'') is a large-scale multi-object video segmentation dataset consisting of $4453$ long-video sequences with around $200$k annotations sparsely provided at every $5^{\text{th}}$ frame. In particular, YTVOS18 is designed as a benchmark for semi-supervised VOS tasks ({\em i.e.,~}first-frame groundtruth provided). As a result, some sequences would be not suitable for motion segmentation evaluation due to problems such as {stationary objects}, unannotated moving objects, {\em etc}. Instead, we select a subset of YTVOS18 to form a new multi-object motion segmentation benchmark, named YouTubeVOS2018-motion  (short form ``YTVOS18-m'').

\subsection{YouTubeVOS2018-Motion Dataset}
\label{supsubsec:datasets_ytvos}

{To curate a new evaluation benchmark for motion segmentation (where object discovery heavily relies on motion information), we manually select $120$ sequences, each containing up to three \textbf{predominantly moving} objects, from the YTVOS18 training split. Note, this dataset is only used for evaluation purposes.}
Compared to the DAVIS sequences, the selected YTVOS sequences comprise more realistic situations with variable motion patterns including a mixture of partial motion, background distraction, object occlusion and interactions, {\em etc}. 
We summarise our selection criteria into three major aspects regarding appearance, annotations and object motion, as outlined below.

\vspace{5pt}
{\noindent \em Selection based on Appearance. } The selected sequence should correspond to a smooth and clean video clip. In particular, we filter out sequences that contain noticeable captions, watermarks and margins. Furthermore, videos with abrupt transitions in scenes are not adopted.

\vspace{5pt}
{\noindent \em Selection based on Annotations. } All moving objects should be annotated. This is noted since in some YTVOS18 sequences, annotations are provided only for a subset of moving objects, as it is intended for semi-supervised tracking purposes.

\vspace{5pt}
{\noindent \em Selection based on Object Motion. } To identify sequences suitable for the motion segmentation tasks, we aim to avoid three particular scenarios, from which little object information could be extracted based on motion information: (i) \textbf{common motion}. This follows the same principle as the curation of DAVIS17-m dataset~\cite{Xie22}. Instead of re-annotations, we directly discard sequences with common motions; (ii) \textbf{partial motion}. If the maximum proportion of object parts that move is below a certain threshold ({\em e.g.,} $<30\%$), the sequence would be filtered out as an extreme case of partial motion; (iii) \textbf{stationary frames}. Following a similar definition to the previous criterion, if the sequence contains only a minority of frames with moving objects ({\em e.g.,} $<25\%$), it would not be selected.
\\\\
\noindent
{In summary, YTVOS18-m is introduced as a motion segmentation benchmark, featuring \textbf{multi-object} sequences characterized by \textbf{dominant object movements}. This stands in contrast to the original YTVOS18 dataset, as YTVOS18-m  sequences are curated to support motion-based object discovery, offering a standardised benchmark for evaluating motion segmentation methods. Samples of YTVOS18-m sequences are given in~\Cref{supfig:multivos,supfig:vis-sam,supfig:vis-alpha}.}

\clearpage
\clearpage
\section{Ablation Study}
\label{supsec:ablation}

{In this section, we present comprehensive ablation studies to validate our model performance on multi-object benchmarks. 
Specifically, in Sec.~\ref{supsubsec:ablation-syntrn}, 
we focus on the synthetic pre-training stage, verifying hyperparameter choices and key design elements of our proposed models.
Sec.~\ref{supsubsec:ablation-ssa} provides detailed results at various stages of self-supervised adaptation. This includes self-supervised training of the mask corrector and the adaptation of the flow segmentation model. 
In Sec.~\ref{supsubsec:ablation-invest}, we investigate how our self-supervised training scheme effectively mitigates the Sim2Real disparity introduced by the upsampling process.} 
\revised{Sec.~\ref{supsubsec:sam} demonstrates that SAM-related models can be complementary with our model to achieve further performance improvement.}

\vspace{-0.1cm}
\subsection{{Pre-training on Synthetic Videos}}
\label{supsubsec:ablation-syntrn}
{In Sec.~\ref{supsubsubsec:ablation-selection}, we report the results of our synthetically-trained {\textit{mask selector}} and compare it with other selection mechanisms. Sec.~\ref{supsubsubsec:ablation-correction} provides the comparison of various model configurations for the {\textit{mask corrector}} during the synthetic pre-training stage.}

\vspace{-0.2cm}
\subsubsection{{Synthetically-Trained Mask Selection.}}
\label{supsubsubsec:ablation-selection}
To evaluate the selection accuracy, we propose a metric that assesses the ability of the selector to identify $t = 10$ high-quality 
{exemplar masks per object while avoiding the problematic ones.}
Specifically, for each object, we rank the frames in the sequence based on the IoU values between the flow-predicted masks and the groundtruth segmentation masks, from which the top $10$ frames and the bottom $10$ frames are extracted to form a set of $20$ frames. The selection accuracy is then measured by the Average Precision (AP) of selecting the top $10$ frames from this set.

\begin{table}[!htb]
  \centering
  \small
  \setlength\tabcolsep{8pt}
  \vspace{-0.2cm}
  \begin{tabular}{lc}  
    \toprule
    Selection mechanism  & DAVIS17-m~~AP $\uparrow$  \\
    \midrule
    Random  &  $0.568$   \\
    Propagation-based~\cite{Xie22}  & $0.830$   \\
    \midrule
    \textbf{Ours} -- selector (score) & $0.832$  \\ 
    \textbf{Ours} -- selector (error map)  & $\textbf{0.893}$   \\     
    \bottomrule
  \end{tabular}
  \vspace{0.2cm}
  \caption{\textbf{Comparison across different selection mechanisms.} ``AP'' denotes the Average Precision of selecting $10$ frames with the highest IoU values. ``score'' and ``error map'' correspond to mask selectors trained to directly predict confidence scores and error maps, respectively.}
  \label{tab:ablation-selector}
  \vspace{-0.9cm}
\end{table}

\vspace{3pt}
\par{\noindent \textit{Comparison with Other Selection Mechanism. }}
As shown in Table~\ref{tab:ablation-selector}, our synthetically-trained mask selector outperforms random selection and the method proposed in~\cite{Xie22}, where the latter approach relies on framewise mask propagation using dense DINO features.

\vspace{3pt}
\par{\noindent \textit{Error Map Prediction. }} In addition to the default error map prediction, we explore an alternative approach which trains the selector to directly output confidence scores for each {mask}. However, in Table~\ref{tab:ablation-selector}, this score-based model exhibits a degraded performance compared to the error map prediction, which verifies the effectiveness of predicting error maps as an intermediate step in our design.

\subsubsection{{Synthetically-Trained Mask Correction.}}
\label{supsubsubsec:ablation-correction}
As shown in Table~\ref{suptab:corrector}, despite being trained only synthetically, our proposed mask corrector significantly improves the performance on real-video benchmarks compared to the flow-predicted mask inputs. 

\begin{table}[!htb]
\vspace{-0.6cm}
  \centering
  \small
  \setlength\tabcolsep{6pt}
  \begin{tabular}{ccc}  
    \toprule
    \multirow{2}{*}{\shortstack{Corrector \\ syn-trn} } & \multirow{2}{*}{\shortstack{YTVOS18-m \\ $\mathcal{J} \uparrow$}} & \multirow{2}{*}{\shortstack{DAVIS17-m \\ $\mathcal{J} \uparrow$}}   \\
     &  & \\
    \midrule
    \xmark & $46.5$ & $54.5$   \\
    $\checkmark$ & {\hspace{0.7cm} $\textbf{60.9} {\scriptstyle \; \textcolor{light-red}{+14.4}}$} &  {\hspace{0.55cm} $\textbf{60.0} {\scriptstyle \; \textcolor{light-red}{+5.5}}$}   \\  
    \bottomrule
  \end{tabular}
  \vspace{0.2cm}
  \caption{\textbf{Effect of the synthetically-trained mask corrector}. {``Corrector syn-trn'' is short for synthetically-trained mask corrector.} The two rows correspond to the performance before ({\em i.e.,~}flow-predicted masks) and after the mask correction, respectively.}
  \label{suptab:corrector}
  \vspace{-0.9cm}
\end{table}

\vspace{1pt}
\par{\noindent \textit{Architecture Choice. }}
Table~\ref{suptab:corrector-arch} further introduces two variants of our mask corrector model (first two rows), both with degraded overall performance. 
The former directly upsamples transparency alpha masks generated during feature reconstruction, rather than extracts cross-attention maps from the transformer module. The latter discards the notion of object-centric queries and instead establishes framewise dense spatial correlations in the transformer module. 

\begin{table}[!htb]
\vspace{-0.5cm}
  \centering
  \small
  \setlength\tabcolsep{6pt}
  \begin{tabular}{cccc}  
    \toprule
    \multirow{2}{*}{Attn-map} & \multirow{2}{*}{\shortstack{Object- \\ centric}} & \multirow{2}{*}{\shortstack{YTVOS18-m \\ $\mathcal{J} \uparrow$}} & \multirow{2}{*}{\shortstack{DAVIS17-m \\ $\mathcal{J} \uparrow$}}   \\
     & & & \\
    \midrule
    \xmark & $\checkmark$  & $54.6$  &  $55.8$   \\ 
    - & \xmark & $58.8$  &  $59.2$  \\
    \midrule
    $\checkmark$ & $\checkmark$ & $\textbf{60.9}$ &  $\textbf{60.0}$    \\  
    \bottomrule
  \end{tabular}
  \vspace{0.2cm}
  \caption{\textbf{Architecture choices for the synthetically-trained mask corrector.}}
  \label{suptab:corrector-arch}
  \vspace{-0.9cm}
\end{table}

\vspace{3pt}
\par{\noindent \textit{Number of Frames During Training. }} 
{We also adjust the number of target frames and selected exemplar masks during synthetic training of the mask corrector.}
As demonstrated in Table~\ref{suptab:corrector-T} and Table~\ref{suptab:corrector-t}, the performance remains relatively consistent when varying the number of frames within the specified ranges. 
{This confirms the validity of our choices of $T=7$ target frames and $t=10$ exemplar masks per object.}

{\vspace{0.3cm}
\noindent\begin{minipage}[htbp]{0.48\textwidth}%
\resizebox{\textwidth}{!}{
  \centering
  \small
  \begin{tabular}{ccc} 
    \toprule
    \,\multirow{2}{*}{\shortstack{$\#$ of target \\frames $T$}} \,& \, \multirow{2}{*}{\shortstack{YTVOS18-m \\ $\mathcal{J} \uparrow$}} \, & \, \multirow{2}{*}{\shortstack{DAVIS17-m \\ $\mathcal{J} \uparrow$}} \, \\
     & & \\
    \midrule
    $5$ & $60.6$  &  $60.1$  \\
    $7$ & $60.9$  &  $60.0$   \\ 
    $10$ & $60.9$  &  $59.9$  \\
    \bottomrule
  \end{tabular}}
  \vspace{-0.25cm}
  \captionof{table}{\textbf{Number of Target Frames $T$ During Training.} Results are shown for synthetically-trained mask correctors. The number of {exemplar masks for each object} is fixed at $t = 10$.}
  \label{suptab:corrector-T}
\end{minipage}
\hspace{0.4cm}
\begin{minipage}[htbp]{0.48\textwidth}%
\resizebox{\textwidth}{!}{
  \centering
  \small 
  \begin{tabular}{ccc}  
    \toprule 
     \,\multirow{2}{*}{\shortstack{$\#$ of exemplar \\{masks} $t$}} \,&\, \multirow{2}{*}{\shortstack{YTVOS18-m \\ $\mathcal{J} \uparrow$}} \,& \,\multirow{2}{*}{\shortstack{DAVIS17-m \\ $\mathcal{J} \uparrow$}} \,  \\
     & & \\
    \midrule
    $5$ & $60.3$  &  $59.8$  \\
    $10$ & $60.9$  &  $60.0$  \\ 
    $15$ & $61.2$  &  $59.7$  \\
    \bottomrule
  \end{tabular}}
  \vspace{-0.25cm}
  \captionof{table}{{\textbf{Number of exemplar masks (per object) $t$ during training.}} Results are shown for synthetically-trained mask correctors. The number of target frames is fixed at $T = 7$.}
  \label{suptab:corrector-t}
\end{minipage}
\vspace{0.cm}
}

\vspace{3pt}
\par{\noindent {\textit{Number of Background Queries. }}} {Table~\ref{suptab:bgquery} verifies the choice of adopting multiple background queries ($K_{bg} = 4$), which results in a slight performance boost compared to the single query case. This improvement stems from the decomposition of complicated background information into multiple semantically meaningful regions.}

{Benefiting from this idea, we can also infer different background semantic regions, as will be shown in Figure~\ref{supfig:vis-alpha}.}

\begin{table}[!htb]
\vspace{-0.5cm}
  \centering
  \small
  \setlength\tabcolsep{6pt}
  \begin{tabular}{cc}  
    \toprule 
    \multirow{2}{*}{\shortstack{Number of background \\ queries $K_{bg}$}}  & \multirow{2}{*}{\shortstack{DAVIS17-m \\ $\mathcal{J} \uparrow$}}   \\
     &  \\
    \midrule
    $1$ & $59.3$   \\
    $4$ & $\textbf{60.0}$  \\ 
    \bottomrule
  \end{tabular}
  \vspace{0.25cm}
  \caption{{\textbf{Number of background queries.} Results are shown for synthetically-trained mask correctors.}}
  \label{suptab:bgquery}
  \vspace{-0.9cm}
\end{table}

\vspace{3pt}
\par{\noindent {\textit{Choice of Self-Supervised Features. }}} {We compare across various self-supervised visual features, including DINOv2 (ViT-S/14)~\cite{oquab2023dinov2} and MAE (ViT-B/16)~\cite{He_2022_CVPR}. To ensure a fair comparison, we resize the input images so that the resulting bottleneck resolutions align with the default DINO features. Nevertheless, as indicated in Table~\ref{suptab:ssf}, incorporating other self-supervised visual features does not lead to performance improvements.}

\begin{table}[!htb]
\vspace{-0.5cm}
  \centering
  \small
  \setlength\tabcolsep{6pt}
  \begin{tabular}{lc}  
    \toprule 
    Self-supervised features & DAVIS17-m $\mathcal{J} \uparrow$   \\
    \midrule
    MAE~\cite{He_2022_CVPR} & $55.1$  \\
    DINO~\cite{caron2021emerging} & $\textbf{60.0}$  \\
    DINOv2~\cite{oquab2023dinov2} & $59.2$  \\ 
    \bottomrule
  \end{tabular}
  \vspace{0.25cm}
  \caption{{\textbf{Self-supervised features adopted in the synthetically-trained mask corrector.}}}
  \label{suptab:ssf}
  \vspace{-0.9cm}
\end{table}

\vspace{3pt}
\par{\noindent {\textit{Choice of Optical Flow Methods. }} {The choice of flow estimation technique could influence the quality of flow-predicted segmentation, which can in turn affect the performance of refined masks. This is highlighted in Table~\ref{suptab:flow}, where the default RAFT approach outperforms other optical flow methods.}

\begin{table}[!htb]
  \centering
  \small
  \setlength\tabcolsep{6pt}
  \begin{tabular}{lcc}  
    \toprule 
    \multirow{2}{*}{\shortstack{Optical flow \\ method}}  & \multicolumn{2}{c}{DAVIS17-m $\mathcal{J} \uparrow$}   \\
    \cmidrule(r){2-3}
     & OCLR-flow & OCLR-flow + \textbf{Our} refinements\\
    \midrule
    ARFlow~\cite{liu2020learning} & $39.5$ &  $53.2$ \\
    MaskFlownet~\cite{zhao2020maskflownet} & $49.0$  &  $57.1$ \\ 
    RAFT~\cite{Teed20} & $\textbf{54.5}$ &  $\textbf{60.0}$\\ 
    \bottomrule
  \end{tabular}
  \vspace{0.25cm}
  \caption{{\textbf{Optical flow methods}. {In the column heading, ``Our refinements'' indicates that we refine the flow-predicted masks obtained by OCLR-flow, using the synthetically-trained selection-correction process.}}}
  \label{suptab:flow}
  \vspace{-0.9cm}
\end{table}

\vspace{3pt}
\par{\noindent {\textit{Choice of Flow Segmentation Models. }}} {The choice of flow-based segmentation models could also affect the quality of refined masks. As demonstrated in Table~\ref{suptab:flowseg}, our appearance-based refinements noticeably enhance the performance of flow-predicted proposals from various flow-based models, among which OCLR-flow leads to the top performance.} 

\begin{table}[!htb]
\vspace{-0.3cm}
  \centering
  \small
  \setlength\tabcolsep{6pt}
  \begin{tabular}{lcc}  
    \toprule 
    \multirow{3}{*}{\shortstack{Flow \\ segmentation \\ model}}  & \multicolumn{2}{c}{DAVIS17-m $\mathcal{J} \uparrow$}   \\
    \cmidrule(r){2-3}
     & \multirow{2}{*}{Flow-predicted masks} & \multirow{2}{*}{\shortstack{Flow-predicted masks \\ + \textbf{Our} refinements}}\\
     & & \\
    \midrule
    MG~\cite{Yang21a} & $38.4$ &  $48.4$ \\
    MG-sup.~\cite{Yang21a} & $44.9$  &  $59.2$ \\ 
    Mask R-CNN~\cite{He_2017_ICCV} & $52.0$ &  $57.4$ \\ 
    OCLR-flow~\cite{Xie22} & $\textbf{54.5}$ &  $\textbf{60.0}$ \\
    \bottomrule 
  \end{tabular}
  \vspace{0.25cm}
  \caption{\textbf{Flow segmentation models}. {In the column heading, ``Our refinements'' indicates that we refine flow-predicted masks obtained by different flow segmentation models, using the synthetically-trained selection-correction process.}}
  \label{suptab:flowseg}
  \vspace{-1.3cm}
\end{table}

\subsubsection{Repeating (Multiple Run) Experiments.}

We repeat the synthetic training for both the mask selector and corrector using the same hyperparameter settings, with the results summarised in Table~\ref{suptab:repeat-selector} and Table~\ref{suptab:repeat-corrector}. Notably, the performance differences between the repeated experiments are generally minimal, which confirms the consistency and reliability of our results.

{\vspace{0.3cm}
\noindent\begin{minipage}[htbp]{0.48\textwidth}%
  \centering
  \small
  \begin{tabular}{cc}  
    \toprule
    \setlength\tabcolsep{10pt}
    Selector exp.  & DAVIS17-m~~AP $\uparrow$    \\
    \midrule
    1 & $0.893$  \\ 
    2  & $0.886$  \\ 
    3  & $0.897$  \\ 
    \midrule
    Mean & $0.892$  \\ 
    Std. & $\pm 0.005$ \\ 
    \bottomrule
  \end{tabular}
  \vspace{-0.1cm}
  \captionof{table}{\textbf{Repeating experiments for the synthetically-trained mask selector.}}
  \label{suptab:repeat-selector}
\end{minipage}
\hspace{0.4cm}
\begin{minipage}[htbp]{0.48\textwidth}%
  \centering
  \small 
  \begin{tabular}{ccc} 
    \toprule 
    \setlength\tabcolsep{10pt}
    \multirow{2}{*}{\shortstack{Corrector \\ exp.}}  &
    \multirow{2}{*}{\shortstack{YTVOS18-m \\ $\mathcal{J} \uparrow$}} & \multirow{2}{*}{\shortstack{DAVIS17-m \\ $\mathcal{J} \uparrow$}}   \\
     & & \\
    \midrule
    1 & $60.7$  &  $59.8$  \\
    2 & $60.9$  &  $60.0$  \\ 
    3 & $60.3$  &  $60.3$  \\
    \midrule
    Mean & $60.6$  &  $60.0$  \\
    Std. & $\pm 0.2$  &  $\pm 0.2$  \\
    \bottomrule
  \end{tabular}
  \vspace{-0.1cm}
   \captionof{table}{\textbf{Repeating experiments for the synthetically-trained mask corrector.}}
   \label{suptab:repeat-corrector}
\end{minipage}
\vspace{0cm}
}

\subsection{{Self-Supervised Adaptation to Real-World Videos}}
\label{supsubsec:ablation-ssa}

\par{\noindent {\textbf{Adapting the Mask Correction Model.}}} {We adopt a self-supervised adaptation process for the synthetically-trained mask corrector to reduce the Sim2Real disparity. This results in a significant performance boost as shown in Table~\ref{suptab:adap}. The designed adaptation is conducted for each target real-world video dataset. Nevertheless, we found that the additional per-sequence self-supervised training (i.e., test-time training) would further improve the refined mask quality.} 

\begin{table*}[!ht]
\vspace{-0.cm}
  \centering
  \small
  \setlength\tabcolsep{8pt}
  \resizebox{\textwidth}{!}{
  \begin{tabular}{ccccccc}  
    \toprule
   \multirow{2}{*}{ \shortstack{Corrector \\ adap.}}  & \multirow{2}{*}{\shortstack{Flow model \\ adap.}} & \multirow{2}{*}{ \shortstack{Corrector \\ adap. (seq.)}}  & \multirow{2}{*}{CRF} &  \multirow{2}{*}{\shortstack{Inference \\ time / s}} & \multirow{2}{*}{\shortstack{YTVOS18-m \\ $\mathcal{J} \uparrow$}} & \multirow{2}{*}{\shortstack{DAVIS17-m \\ $\mathcal{J} \uparrow$}}   \\
    & & & & & &  \\
    \midrule
    \xmark & \xmark & \xmark  & \xmark & $0.098$ & {$60.9$} &  {$60.0$}    \\ 
    $\checkmark$ & \xmark & \xmark  & \xmark & $0.098$ &{\hspace{0.55cm} $64.6 {\scriptstyle \; \textcolor{light-red}{+3.7}}$} &  {\hspace{0.55cm} $63.2 {\scriptstyle \; \textcolor{light-red}{+3.2}}$}    \\ 
    $\checkmark$ & $\checkmark$ & \xmark  & \xmark & $0.098$ &{\hspace{0.55cm} $66.6 {\scriptstyle \; \textcolor{light-red}{+2.0}}$} &  {\hspace{0.55cm} $65.4 {\scriptstyle \; \textcolor{light-red}{+2.2}}$}  \\ 
    $\checkmark$ & $\checkmark$  & $\checkmark$  & \xmark & $2.030$ &{\hspace{0.55cm} $66.9 {\scriptstyle \; \textcolor{light-red}{+0.3}}$} &  {\hspace{0.55cm} $66.3 {\scriptstyle \; \textcolor{light-red}{+0.9}}$}   \\ 
    $\checkmark$ & $\checkmark$ & $\checkmark$ & $\checkmark$  & $5.600$ & {\hspace{0.55cm} $\textbf{67.1} {\scriptstyle \; \textcolor{light-red}{+0.2}}$} &  {\hspace{0.55cm} $\textbf{67.0} {\scriptstyle \; \textcolor{light-red}{+0.7}}$}   \\ 
    \bottomrule
  \end{tabular}}
  \vspace{0.2cm}
  \caption{{\textbf{Self-supervised adaptation stages.} {``adap.'' is short for adaptation conducted at the dataset level.} ``Corrector adap. (seq.)'' represents per-sequence test-time adaptation exclusively conducted for the mask corrector. Inference time denotes the average time taken to obtain one frame of the segmentation result, starting from the RGB and optical flow inputs.}}
  \label{suptab:adap}
  \vspace{-0.7cm}
\end{table*}

\vspace{3pt}
\par{\noindent {\textbf{Adapting the Flow Segmentation Model. }}} {After the appearance-based refinement, we perform the flow segmentation model adaptation by utilising the refined masks as pseudo-GT to fine-tune the synthetically-trained motion segmentation model ({\em i.e.,~}OCLR-flow) on real-world videos. The updated flow-predicted masks from the adapted OCLR-flow model are then refined through the same appearance-based selection-correction procedure, leading to the results shown in Table~\ref{suptab:adap}. 
The overall performance can be further improved by applying CRF as a post-processing step.}

\vspace{5pt}
\par{\noindent \textbf{Feature Reconstruction by the Mask Corrector. }}
Table~\ref{suptab:correctorTTT-recon} highlights the benefits of the auxiliary feature reconstruction process during the self-supervised training of the mask corrector, which improves overall results by encouraging the model to cluster the video scene into spatially compact regions.

\begin{table}[!htb]
  \centering
  \small
  \setlength\tabcolsep{6pt}
  \vspace{-0.3cm}
  \begin{tabular}{ccc}  
    \toprule 
    \multirow{2}{*}{\shortstack{Feature \\ recon.}} & \multirow{2}{*}{\shortstack{YTVOS18-m \\ $\mathcal{J} \uparrow$}} & \multirow{2}{*}{\shortstack{DAVIS17-m \\ $\mathcal{J} \uparrow$}}   \\
     &  & \\
    \midrule
    \xmark & $63.7$  &  $62.6$    \\ 
    $\checkmark$ & $\textbf{65.0}$ &  $\textbf{63.8}$  \\ 
    \bottomrule
  \end{tabular}
  \vspace{0.25cm}
  \caption{\textbf{Feature reconstruction process during the self-supervised adaptation of the mask corrector.} The adaptation is conducted at both dataset- and sequence-level.}
  \label{suptab:correctorTTT-recon}
  \vspace{-0.9cm}
\end{table}

\subsection{{Supervised Adaptation: An Investigation}}
\label{supsubsec:ablation-invest}
Adopting self-supervised DINO features helps to reduce the Sim2Real disparity within the synthetically-trained mask corrector, especially for the transformer module. However, the upsampling process ({\em i.e.,~}the CNN encoder and decoder) could still suffer from the adaptation problem, as it directly takes in raw RGB frames and flow-predicted masks as inputs.

To assess the Sim2Real disparity introduced in the upsampling process, we {consider a \textbf{supervised} adaptation strategy. This involves fine-tuning} the upsampling module using a large number of real video sequences with human annotations, specifically the DAVIS17 and YTVOS18 training sets. It is important to note that this study is conducted \textbf{for investigation purposes only} and does \textbf{not} contribute to our final results. The first two rows in Table~\ref{suptab:upsampleFT} demonstrate the effectiveness of fine-tuning the upsampling module for the synthetically-trained mask corrector. Nevertheless, the benefits of fine-tuning diminish after the {self-supervised} adaptation process, which suggests that our proposed {self-supervised} training scheme partially complements the fine-tuning process, therefore contributing to the Sim2Real generalisation of the upsampling module.

\begin{table}[!htb]
\vspace{-0.2cm}
  \centering
  \small
  \setlength\tabcolsep{6pt}
  \begin{tabular}{cccc}  
    \toprule
    \multirow{2}{*}{\shortstack{Corrector \\ adap. (w/ seq.)}} & \multirow{2}{*}{\shortstack{Upsamp. \\ FT}} & \multirow{2}{*}{\shortstack{YTVOS18-m \\ $\mathcal{J} \uparrow$}} & \multirow{2}{*}{\shortstack{DAVIS17-m \\ $\mathcal{J} \uparrow$}}   \\
     & & & \\
    \midrule
    \xmark & \xmark & $60.9$ & $60.0$   \\
    \xmark & $\checkmark$ & {\hspace{0.55cm} $63.2 {\scriptstyle \; \textcolor{light-red}{+2.3}}$} & {\hspace{0.55cm} $61.5 {\scriptstyle \; \textcolor{light-red}{+1.5}}$}   \\
    \midrule
    $\checkmark$ & \xmark & $65.0$ & $63.8$   \\
    $\checkmark$ & $\checkmark$ & {\hspace{0.55cm} $66.7 {\scriptstyle \; \textcolor{light-red}{+1.7}}$} & {\hspace{0.55cm} $64.1 {\scriptstyle \; \textcolor{light-red}{+0.3}}$}   \\
    \bottomrule
  \end{tabular}
  \vspace{0.25cm}
  \caption{{\textbf{Supervised fine-tuning of the upsampling module.} ``Corrector adap. (w/ seq.)'' refers to \textit{self-supervised training} of the mask corrector at both dataset- and sequence-level, and ``Upsamp. FT'' stands for \textit{supervised} fine-tuning of the upsampling module on human-annotated real video sequences.}} 
  \label{suptab:upsampleFT}
  \vspace{-0.9cm}
\end{table}

\subsection{Using SAM for Further Refinement}
\label{supsubsec:sam}

\vspace{5pt}
\par{\noindent \textbf{Effect of Different Post-Processing Steps. }} \revised{Table~\ref{suptab:post-process} compares across different post-processing approaches, for example, CRF, SAM, and HQSAM. 
Note that \emph{our main framework} is human-label-free and only adopts CRF as the post-processing step. Furthermore, we demonstrate a complementary performance boost when applying SAM or HQSAM~\cite{sam_hq} (both trained with human annotations) as additional refinement steps.}

\begin{table}[!htb]
  \centering
  \small
  \setlength\tabcolsep{5pt}
  \begin{tabular}{ccccc}  
    \toprule
    \multirow{2}{*}{\shortstack{Human \\ Anno.}} & \multirow{2}{*}{CRF} & \multirow{2}{*}{\shortstack{(HQ)SAM-based \\post-processing} } & \multirow{2}{*}{\shortstack{YTVOS18-m \\ $\mathcal{J} \uparrow$}} & \multirow{2}{*}{\shortstack{DAVIS17-m \\ $\mathcal{J} \uparrow$}}   \\
     &  & \\
    \midrule
    \xmark & \xmark & \xmark & $66.9$ & $66.3$   \\
    \xmark & $\checkmark$ & \xmark & $67.1$ & $67.0$   \\
    \midrule
    $\checkmark$ & \xmark & SAM $\vert$ HQSAM & $70.6$ $\vert$ $70.9$   & $70.4$ $\vert$ $71.2$   \\
    $\checkmark$ & $\checkmark$ & SAM $\vert$ HQSAM & $71.1$ $\vert$ $71.1$ & $70.9$ $\vert$ $71.3$   \\
    \bottomrule
  \end{tabular}
  \vspace{0.15cm}
  \caption{\revised{\textbf{Effect of different post-processing steps.}}}
  \label{suptab:post-process}
  \vspace{-0.45cm}
\end{table}

\vspace{5pt}
\par{\noindent \textbf{Comparison with OCLR-TTA + SAM.}} 
\revised{We apply SAM-based refinement on both our method and OCLR-TTA (previous SoTA). As shown in~Table~\ref{suptab:oclrttasam}, our method with SAM (Ours + SAM) outperforms OCLR-TTA with SAM (OCLR-TTA + SAM), particularly on the challenging YTVOS18-m dataset.}

\begin{table}[!htb]
  \centering
  \small
  \setlength\tabcolsep{5pt}
  \begin{tabular}{cccc}  
    \toprule
    \multirow{2}{*}{Method} & \multirow{2}{*}{\shortstack{Human \\ Anno.}} & \multirow{2}{*}{\shortstack{YTVOS18-m \\ $\mathcal{J} \uparrow$}} & \multirow{2}{*}{\shortstack{DAVIS17-m \\ $\mathcal{J} \uparrow$}}   \\
     &  & \\
    \midrule
    OCLR-TTA + SAM & $\checkmark$ & $62.3$ & $69.4$   \\
    Ours + SAM & $\checkmark$ & $71.1$ & $70.9$  \\
    \bottomrule
  \end{tabular}
  \vspace{0.15cm}
  \caption{\revised{\textbf{Comparison with OCLR-TTA + SAM.}}}
  \vspace{-0.2cm}
  \label{suptab:oclrttasam}
\end{table}
\section{Additional Quantitative Results}
\label{supsec:quantitative}
Table~\ref{suptab:multivos} presents a summary of the quantitative performance on multi-object video segmentation benchmarks. 
Following~\cite{Xie22}, we additionally report the results of three synthetically-supervised models: a supervised version of motion grouping~\cite{Yang21a} (MG-sup.) and two Mask R-CNN~\cite{He_2017_ICCV} models with the first one taking in only optical flows, and the second one considering both RGB and flow inputs. All models are trained using the same synthetic dataset as in our study.

In Table~\ref{suptab:singlevos}, we provide a more comprehensive comparison across different methods for single object video segmentation.

\begin{table*}[!htbp]
\centering
\small
\setlength\tabcolsep{5pt}
  \resizebox{\textwidth}{!}{
  \begin{tabular}{rccccccccccc}  
    \toprule
    {} & \multicolumn{5}{c}{Model settings} & \multicolumn{3}{c}{YTVOS18-m} & \multicolumn{3}{c}{DAVIS17-m}  \\
    \cmidrule(r){2-6}
    \cmidrule(r){7-9}
    \cmidrule(r){10-12}
    Model    & VOS  & H.A. &  RGB &  Flow & Input Res.  & {$\mathcal{J}\&\mathcal{F}$  $\uparrow$} & {$\mathcal{J}$ $\uparrow$} & {$\mathcal{F}$  $\uparrow$} & {$\mathcal{J}\&\mathcal{F}$  $\uparrow$} & {$\mathcal{J}$ $\uparrow$} & {$\mathcal{F}$  $\uparrow$}\\
    \midrule
    $^{\dag}$MG~\cite{Yang21a} & un-sup. & \xmark & \xmark & $\checkmark$ & $128 \times 224$ & $33.3$ & $37.0$ & $29.6$ &  $35.8$ & $38.4$  &  $33.2$    \\
    $^{\dag}$MG-sup.~\cite{Yang21a} & un-sup. & \xmark & \xmark  & $\checkmark$  & $128 \times 224$ & $33.0$ & $39.1$ & $26.9$ &  $39.5$ & $44.9$  &  $34.2$      \\
    $^{\dag}$Mask R-CNN~\cite{He_2017_ICCV} & un-sup. & \xmark & \xmark  & $\checkmark$ & $128 \times 224$ & $41.1$ & $42.9$ & $39.2$  &  $52.2$ & $52.0$  &  $52.3$     \\
    $^{\dag}$Mask R-CNN~\cite{He_2017_ICCV} & un-sup. & \xmark & $\checkmark$  & $\checkmark$ & $128 \times 224$ & $38.8$ & $39.3$ & $38.2$ & $50.4$ & $49.8$ & $50.9$     \\
    OCLR-flow~\cite{Xie22} &  un-sup. & \xmark &  \xmark  &$\checkmark$ & $128 \times 224$ & $45.3$ & $46.5$ & $44.1$ &  $55.1$ &  $54.5$  &  $55.7$     \\
    {Safadoust et al.~\cite{Safadoust23}}  &  un-sup. & \xmark &  $\checkmark$ & \xmark & $128 \times 224$ & $-$ & $-$ & $-$ &  $59.2$ &  $59.3$  &  $59.2$     \\
    OCLR-TTA~\cite{Xie22} & un-sup. & \xmark & $\checkmark$  & $\checkmark$ & $480 \times 854$ & $50.6$ & $52.7$ & $48.6$  &  $64.4$ &  $65.2$  &  $63.6$   \\
    {$^{\dag}$VideoCutLER~\cite{wang2023videocutler}} &  un-sup. & \xmark &  $\checkmark$ & \xmark & $480 \times 854$ & $57.0$ & $59.0$ & $55.1$ &  $57.3$ &  $57.4$  &  $57.2$     \\
    \textbf{Ours} & un-sup. & \xmark & $\checkmark$  & $\checkmark$ & $128 \times 224$ & $\textbf{65.2}$ & $\textbf{67.1}$ & $\textbf{63.2}$ &  $\textbf{66.2}$ &  $\textbf{67.0}$  &  $\textbf{65.4}$   \\
    \midrule
    {OCLR-flow~\cite{Xie22} + SAM~\cite{kirillov2023segany}} &  un-sup. & $\checkmark$ & $\checkmark$  & $\checkmark$ & $576 \times 1024$ & $58.5$ & $57.0$ & $60.0$ &  $64.2$ &  $62.0$  &  $66.4$     \\
    {\textbf{Ours} + SAM~\cite{kirillov2023segany}} &  un-sup. & $\checkmark$ & $\checkmark$  & $\checkmark$ & $576 \times 1024$ & $\textbf{70.6}$ & $\textbf{71.1}$ & $\textbf{70.2}$ &  $\textbf{71.5}$ &  $\textbf{70.9}$  &  $\textbf{72.1}$     \\
    \midrule
    $^{\dag}$CorrFlow~\cite{Lai19} & semi-sup. & \xmark & $\checkmark$  &  \xmark & $480 \times 854$ & $55.2$ & $60.0$ & $50.4$  &  $54.0$ &  $54.2$  &  $53.7$   \\
    $^{\dag}$UVC~\cite{nips19_joint_task} & semi-sup. & \xmark & $\checkmark$  &  \xmark & $480 \times 854$ & $66.9$ & $70.4$ & $63.4$  &  $65.5$ &  $66.2$  &  $64.7$  \\
    $^{\dag}$MAST~\cite{Lai20} & semi-sup.  & \xmark & $\checkmark$  &  \xmark & $480 \times 854$ & $70.5$ & $72.7$ & $68.2$  &  $70.9$ &  $71.0$  &  $70.8$    \\
    $^{\dag}$CRW~\cite{jabri2020walk} & semi-sup. & \xmark & $\checkmark$  & \xmark & $480 \times 854$ & $72.9$ & $75.5$ & $70.4$ & $73.4$ &  $72.9$  &  $74.1$ \\
    $^{\dag}$DINO~\cite{caron2021emerging} & semi-sup. & \xmark & $\checkmark$  &  \xmark & $480 \times 854$ & $\textbf{75.4}$ & $\textbf{76.6}$ & $\textbf{74.3}$  & $\textbf{78.7}$ &  $\textbf{77.7}$  &  $\textbf{79.6}$  \\
    \bottomrule
  \end{tabular}}
  \vspace{0.25cm}
  \caption{\textbf{Quantitative comparison on multi-object video segmentation benchmarks.} ``H.A.'' indicates that human annotations are involved during training.  {$\dag$ denotes that we reproduce the results on the benchmark datasets using the official code released by the original authors.} In the semi-supervised setting, the first-frame ground-truth annotation is provided.}
  \label{suptab:multivos}
  \vspace{-0.9cm}
\end{table*}  

\begin{table*}[!htbp]
   \centering
   \small  
  \setlength\tabcolsep{8pt}
  \resizebox{\textwidth}{!}{
  \begin{tabular}{rcccccccc}  
    \toprule
    {} & \multicolumn{4}{c}{Model settings} & \multirow{2}[2]{*}{\shortstack{DAVIS16 \\ $\mathcal{J} \uparrow$}} & \multirow{2}[2]{*}{\shortstack{STv2 \\ $\mathcal{J} \uparrow$}}  & \multirow{2}[2]{*}{\shortstack{FBMS59 \\ $\mathcal{J} \uparrow$}}  \\
    \cmidrule(r){2-5}
    Model    &  H.A.  & RGB &  Flow  &  Input Res. &  &  &\\
    \midrule
    SAGE~\cite{sage} & \xmark  & $\checkmark$  &$\checkmark$  & $-$ &  $42.6$ &  $57.6$  &  $61.2$     \\
    NLC~\cite{faktor2014videonlc} & \xmark  & $\checkmark$  &$\checkmark$ & $ - $ &  $55.1$ &  $67.2$  &  $51.5$  \\
    CUT~\cite{Keuper15} & \xmark  &  $\checkmark$  &$\checkmark$  & $ - $ & $55.2$ &  $54.3$  &  $57.2$     \\
    FTS~\cite{Papazoglou13} & \xmark &  $\checkmark$  &$\checkmark$  & $ - $ &  $55.8$ &  $47.8$  &  $47.7$   \\
    AMD~\cite{liu2021emergence} & \xmark  &$\checkmark$  & \xmark &  $128 \times 224$  & $57.8$ & $57.0$ & $47.5$ \\
    $^{\ddagger}$SIMO~\cite{Lamdouar21} & \xmark & \xmark  &$\checkmark$ & $128 \times 128$ &  $67.8$ &  $62.0$  &  $-$     \\
    MG~\cite{Yang21a} & \xmark & \xmark  &$\checkmark$ & $128 \times 224$ &  $68.3$ &  $58.6$  &  $53.1$ \\
    EM~\cite{arxiv.2201.02074} & \xmark & \xmark  &$\checkmark$ & $128 \times 224$ & $69.3$ &  $55.5$  &  $57.8$ \\
    CIS~\cite{yang_loquercio_2019} & \xmark & $\checkmark$  &$\checkmark$ & $192 \times 384$ &  $71.5$ &  $62.0$  &  $63.5$   \\
    {SMTC~\cite{yang_loquercio_2019}} & \xmark & $\checkmark$  &\xmark & $256 \times 256$ &  $71.8$ &  $69.3$  &  $68.4$   \\
    $^{\ddagger}$OCLR-flow~\cite{Xie22}& \xmark  & \xmark  &$\checkmark$ & $128 \times 224$ &  $72.1$ &  $67.6$  &  $70.0$ \\
    {Meunier et al.~\cite{10204532}} & \xmark & \xmark  &$\checkmark$ & $128 \times 224$ &  $73.2$ &  $55.0$  &  $59.4$ \\
    {FODVid~\cite{fodvid}} & \xmark  &$\checkmark$  & \xmark & $256 \times 512$ &  $78.7$ &  $-$ & $-$ \\
    DS~\cite{ye2022sprites} & \xmark  &$\checkmark$  &$\checkmark$ & $240 \times 426$ &  $79.1$ &  $72.1$ & $71.8$ \\
    MOD~\cite{ding2022motioninductive} & \xmark  &$\checkmark$  &$\checkmark$ &  $480 \times 854$ &  $79.2$ &  $69.4$ & $66.9$ \\
    DystaB~\cite{Yang_2021_CVPR} & \xmark  &$\checkmark$  &$\checkmark$ & $192 \times 384$ &  $80.0$ &  $74.2$ & $73.2$ \\
    SSL-VOS~\cite{ponimatkin2023sslvos} & \xmark  &$\checkmark$  &$\checkmark$ & $768 \times 768$  &  $80.2$ &  $74.9$ & $70.0$ \\
    GWM~\cite{Choudhury22} & \xmark  &$\checkmark$  & \xmark &  $128 \times 224$  &  $80.7$ &  $78.9$ & $78.4$ \\
    $^{\ddagger}$OCLR-TTA~\cite{Xie22} & \xmark  &$\checkmark$  &$\checkmark$ &  $480 \times 854$ &  $80.9$ &  $72.3$ & $72.7$ \\
    {LOCATE~\cite{LOCATE}} & \xmark  &$\checkmark$  & \xmark &  $480 \times 848$ &  $80.9$ &  $\textbf{79.9}$ & $68.8$ \\
    $^{\ddagger}$\textbf{Ours} & \xmark  &$\checkmark$  &$\checkmark$ &  $128 \times 224$ &  $\textbf{81.1}$ &  $76.6$ & $\textbf{81.9}$ \\
    \midrule
    FSEG~\cite{Jain17} & $\checkmark$ & $\checkmark$  & $\checkmark$ & $ - $  &  $70.7$ &  $61.4$  &  $68.4$ \\
    LVO~\cite{Tokmakov19} & $\checkmark$ & $\checkmark$  & $\checkmark$ & $ - $  &  $75.9$ &  $57.3$  &  $65.1$ \\
    ARP~\cite{song2018pyramidpdb} & $\checkmark$ & $\checkmark$  & $\checkmark$ & $ 473 \times 473 $  &  $76.2$ &  $57.2$  &  $59.8$ \\
    COSNet~\cite{Lu_2019_CVPR}& $\checkmark$ & $\checkmark$  & \xmark  & $473 \times 473$  &  $80.5$ &  $49.7$   &  $75.6$    \\
    {OCLR-flow~\cite{Xie22} + SAM~\cite{kirillov2023segany}} & $\checkmark$ & $\checkmark$  & $\checkmark$ & $576 \times 1024$ & $80.6$ & $71.5$ & $79.2$    \\
    MATNet~\cite{zhou20} & $\checkmark$  & $\checkmark$  &$\checkmark$ & $473 \times 473$ &  $82.4$  &  $50.4$   &  $76.1$    \\
    DystaB~\cite{Yang_2021_CVPR} & $\checkmark$  &$\checkmark$  &$\checkmark$ & $192 \times 384$ &  $82.8$ &  ${74.2}$ & $75.8$ \\
    AMC-Net~\cite{amcnet} & $\checkmark$ & $\checkmark$  & $\checkmark$  & $ 384 \times 384 $ &  $84.5$  &  $-$   &  $76.5$    \\
    TransportNet~\cite{transportnet} & $\checkmark$ & $\checkmark$  & $\checkmark$  & $ 512 \times 512 $ &  $84.5$  &  $-$   &  $78.7$    \\
    PMN~\cite{pmn} & $\checkmark$  & $\checkmark$  & $\checkmark$ & $512 \times 512$ &  ${85.6}$  &  $-$   &  ${77.8}$   \\
    TMO~\cite{tmo} & $\checkmark$  & $\checkmark$  & $\checkmark$ & $512 \times 512$ &  ${85.6}$  &  $-$   &  ${79.9}$   \\
    {\textbf{Ours} + SAM~\cite{kirillov2023segany}} & $\checkmark$ & $\checkmark$  & $\checkmark$ & $576 \times 1024$ & $86.6$ & $\textbf{81.3}$  & $\textbf{85.7}$      \\
    DPA~\cite{cho2023dual} & $\checkmark$  & $\checkmark$  & $\checkmark$ & $512 \times 512$ &  $\textbf{87.1}$  &  $-$   &  ${81.0}$   \\
    \bottomrule
  \end{tabular}}
  \vspace{0.25cm}
  \caption{\textbf{Quantitative comparison on single object video segmentation benchmarks.} ``H.A.'' indicates that human annotations are involved during training. $\ddagger$ corresponds to the methods that rely on human-label-free supervision on synthetic data.}
  \label{suptab:singlevos}
\end{table*}

\clearpage
\clearpage
\section{Additional Qualitative Results}
\label{supsec:qualitative}

\par{\noindent \textbf{Predictions at Intermediate Stages. }} 
Figure~\ref{supfig:vis-stage} provides detailed visualisations of our intermediate predictions, presenting the breakdown into four major stages including flow-based proposal, synthetically-trained mask selection, synthetically-trained mask correction, and self-supervised adaptation, respectively.

\begin{figure*}[hbtp]
    \hspace{0.cm}
    \includegraphics[width=0.98\textwidth]{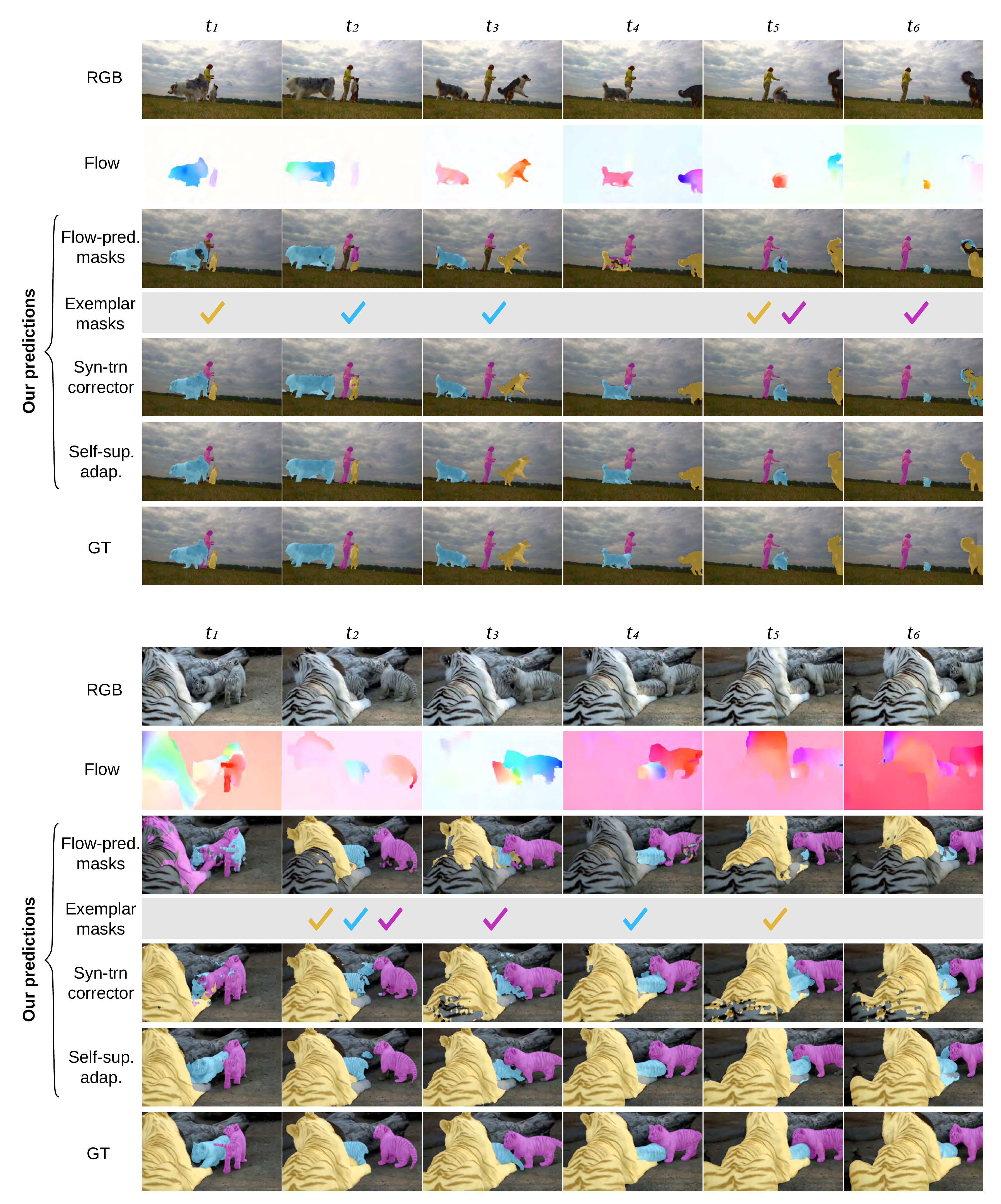}
    \caption{\textbf{Visualisation of our predictions at different stages.} Results are shown for a DAVIS17-m sequence (top) and a YTVOS18-m sequence (bottom). Our predictions consist of four stages from top to bottom: (i) flow-predicted masks; (ii) exemplar masks selected by the mask selector (iii) masks refined by the synthetically-trained mask corrector; (iv) refined masks after the self-supervised adaptation ({\em i.e.,~}the final outputs). In particular, the $\checkmark$ symbols are colour-coded to indicate exemplar masks {selected} for different objects.
    }
    \label{supfig:vis-stage}
\end{figure*}
\vspace{3pt}
\par{\noindent \textbf{Transparency Alpha Masks. }} 
During the feature reconstruction process, transparency alpha masks are predicted at low resolutions, representing the spatial grouping of video scenes as semantic regions ({\em i.e.,~}objects or background components). To recover full resolutions, we upsample these alpha masks using bilinear interpolation. Figure~\ref{supfig:vis-alpha} illustrates predicted alpha masks, demonstrating effective instance-level separation of foreground objects and reasonable clustering of background regions.

\begin{figure*}[hbtp]
    \hspace{0.05cm}
    \includegraphics[width=0.95\textwidth]{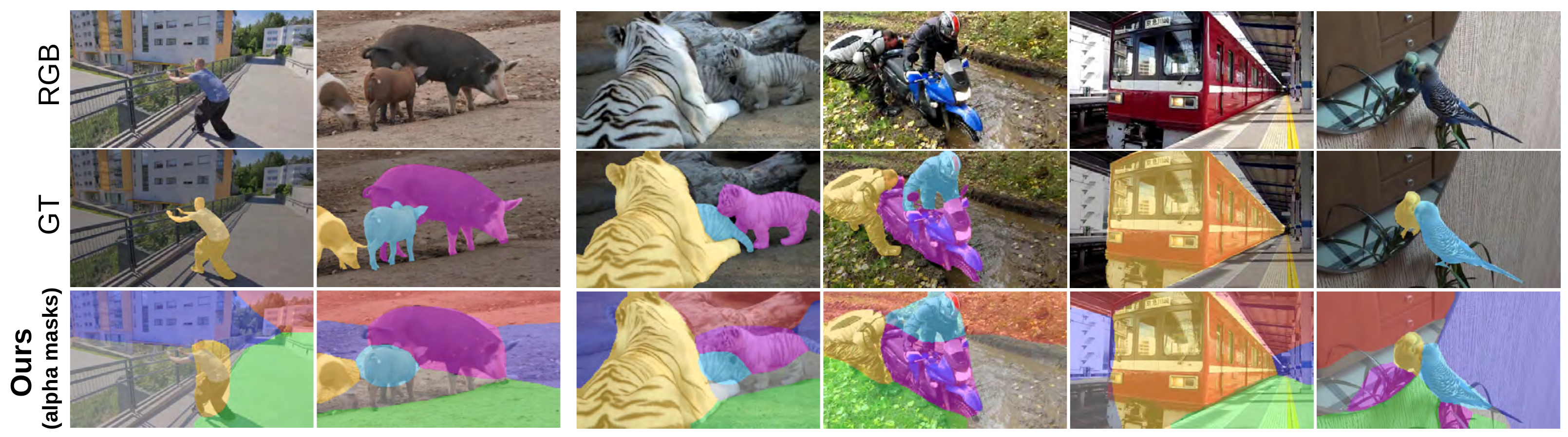}
    \vspace{-0.2cm}
    \caption{\textbf{Visualisation of transparency alpha masks on DAVIS17-m (left) and YTVOS18-m (right).} 
    {These alpha masks, associated with independent queries, are predicted as by-products during the feature reconstruction process and are then upsampled via bilinear interpolation to recover full resolutions. As can be observed, alpha masks segment the entire scene into semantically meaningful regions, which supports our argument that each query corresponds to a distinct object or background region.}}
    \label{supfig:vis-alpha}
    \vspace{-0.55cm}
\end{figure*}

\vspace{0pt}
\par{\noindent {\textbf{Comparison to SAM-Based Refinements. }}}
{To assess the capability of the Segment Anything Model~\cite{kirillov2023segany} as an appearance-based refinement method, we prompt it with segmentation results from OCLR-flow and our method, respectively. The resultant visualisations are provided in Figure~\ref{supfig:vis-sam}, where we make two main observations: (i) While taking the flow-predicted masks ({\em i.e.,~}noisy predictions from OCLR-flow) as inputs, our method (Ours) separates and recovers objects more accurately compared to the SAM-based approach (OCLR-flow + SAM). {This can be attributed to our effective utilization of multi-frame information, in contrast to the per-frame processing by SAM}; (ii) By applying SAM to our outcomes ({\em i.e.,~}Ours + SAM), the boundary details and fine object structures are further refined {marginally}.}

\begin{figure*}[hbtp]
    \hspace{0.cm}
    \includegraphics[width=0.98\textwidth]{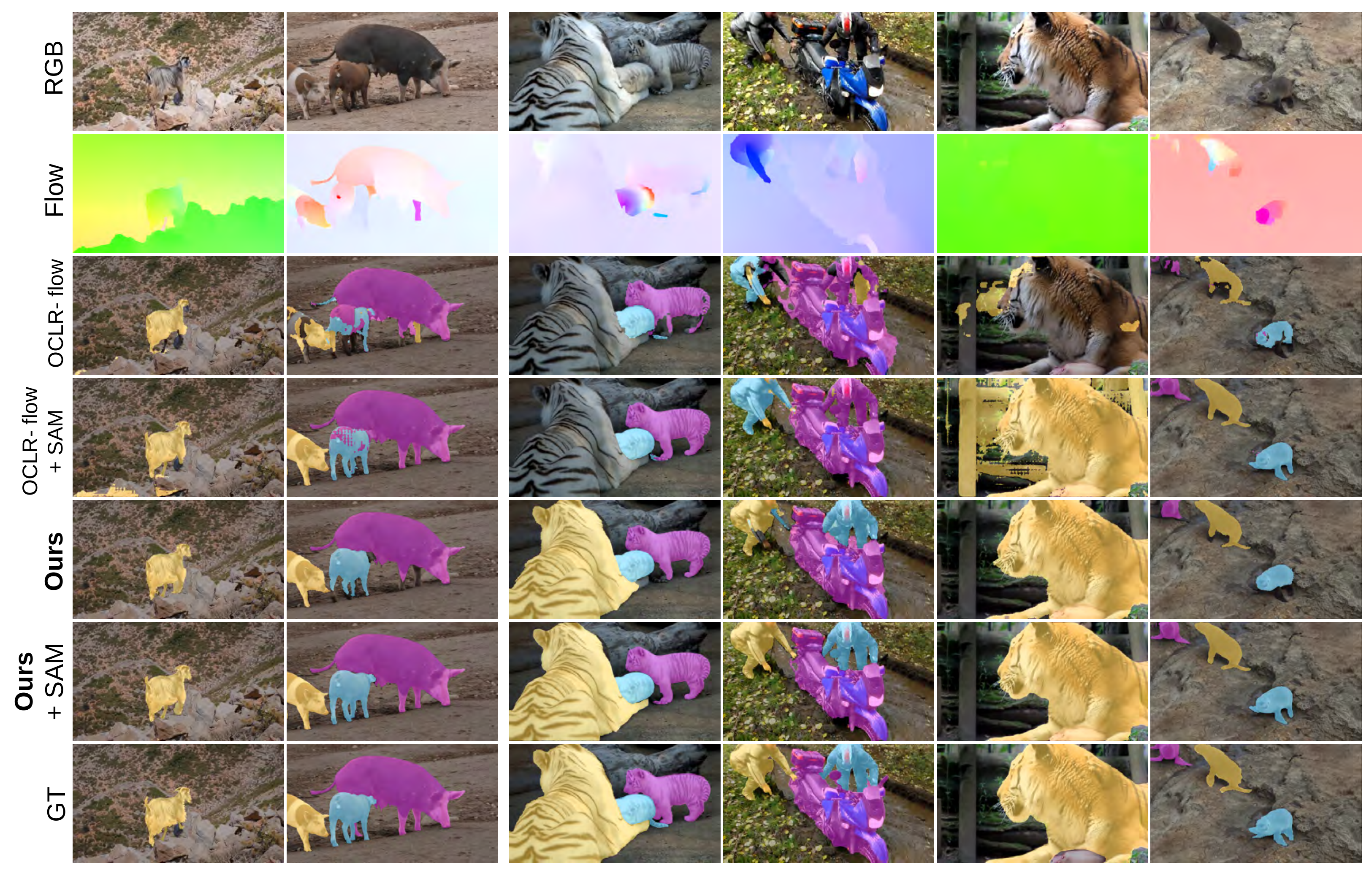}
    \caption{\textbf{Visualisation of the SAM-based refinements applied to OCLR-flow masks and our refined masks.} The results are shown for multi-object video segmentation benchmarks, including DAVIS17-m (left) and YTVOS18-m (right). {``OCLR-flow + SAM'' refers to the results obtained by utilising SAM to refine the OCLR-flow predictions {in a per-frame manner}. As observed in the figure, this approach fails to identify some moving objects, such as the large white tiger in the third column. In contrast, our method (Ours) successfully recovers all object shapes by harnessing temporal correspondence across frames. Furthermore, by applying {per-frame} SAM to refine our results ({\em i.e.,~}Ours + SAM), we observe further boundary enhancements ({\em e.g.,~}around the legs of the purple-masked pig in the second column). This suggests a complementary relationship between our method and the SAM-based refinement.}}
    \label{supfig:vis-sam}
\end{figure*}

\vspace{3pt}
\par{\noindent \textbf{Additional Visualisations. }} 
We provide additional visualisations of multi- and single-object segmentation in Figure~\ref{supfig:multivos} and Figure~\ref{supfig:singlevos}, respectively. In the last column of Figure~\ref{supfig:multivos}, we demonstrate a representative failure case as a result of extremely complex motion patterns including partial motion, object articulation, and interaction.

\begin{figure*}[hbtp]
    \hspace{0.cm}
    \includegraphics[width=0.98\textwidth]{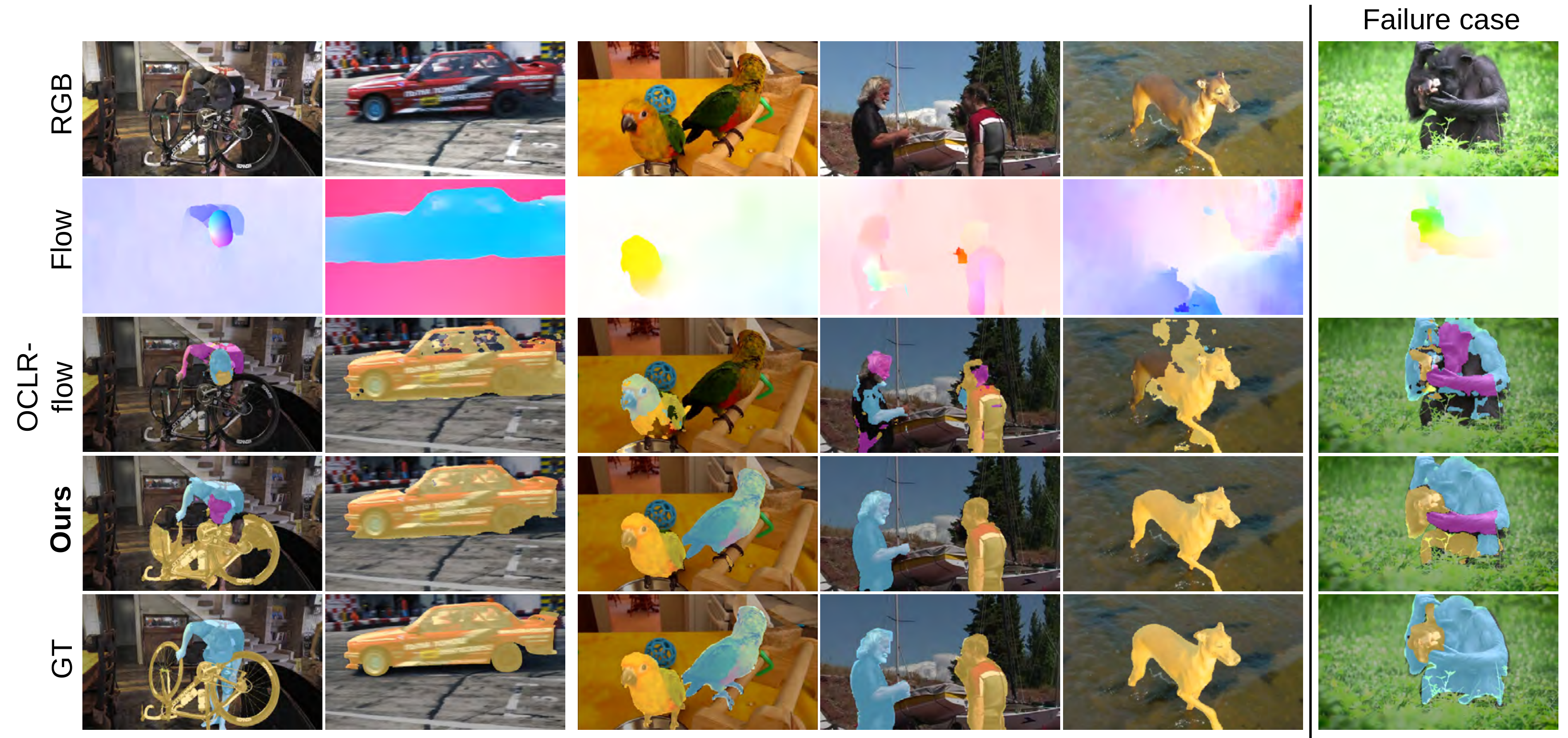}
    \caption{
    {\textbf{Visualisation of multi-object video segmentation on DAVIS17-m (left) and YTVOS18-m (right).} Under complex flow scenarios, the OCLR-flow model struggles to detect objects with subtle movements, such as the stationary bike (first column) and the deer running across the river (fifth column). On the other hand, our method (Ours) accurately recovers these objects by exploiting temporal consistency within the appearance streams. In the last column, we present a challenging case where our approach encounters difficulties arising from complicated motion patterns including partial movement, object articulation, and interaction.}}
    \label{supfig:multivos}
\end{figure*}

\begin{figure*}[hbtp]
    \hspace{0.cm}
    \includegraphics[width=0.98\textwidth]{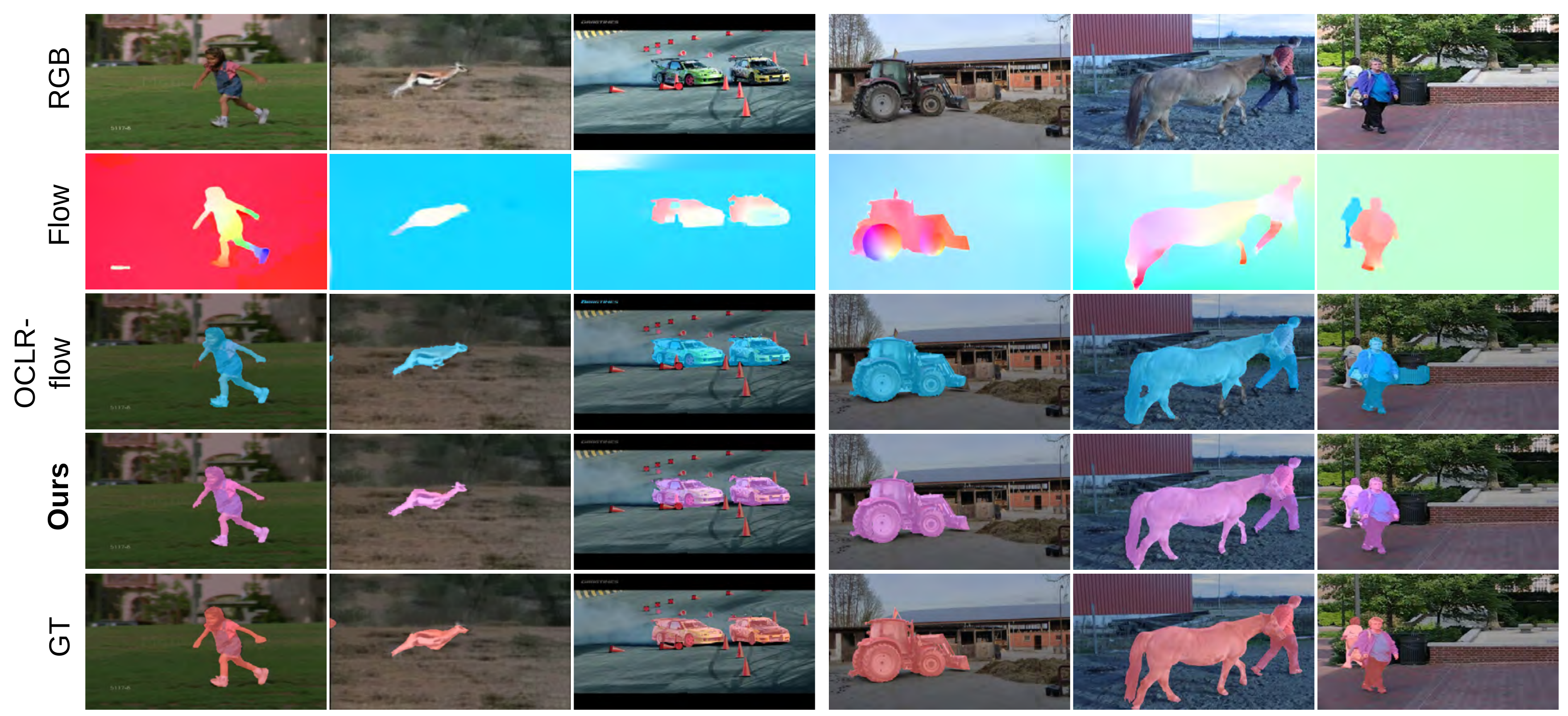}
    \caption{\textbf{Visualisation of single object video segmentation on SegTrackv2 (left) and FBMS-59 (right).} {OCLR-flow correctly captures the majority of objects, and our appearance-based refinement method provides further improvements to the boundary details of segmentation masks.}}
    \label{supfig:singlevos}
\end{figure*}

\vspace{-0.15cm}
\section{Potential Societal Impacts}
\label{appendix:ethic}
\vspace{-0.1cm}
Our goal is a method to discover moving objects in videos and refine their segmentation masks based on appearance information in a class-agnostic manner. The whole process does not rely on training from human annotations, therefore avoiding introducing potential biases from human prior knowledge in the labeling. The synthetic dataset for training is generated by a simulation pipeline (originally introduced in~\cite{Xie22}) that carefully filters out personally identifiable information, and this eliminates the risk of introducing sensitive information and human-related biases. For evaluation, we adopt popular video segmentation benchmarks that follow the CC-BY license with necessary references made. While we have made efforts to mitigate potential negative societal impacts, it is important to note that our method can be adapted to real-world videos without supervision, which could raise the concern of potential misuse by others when applied to inappropriate videos.

\clearpage
\clearpage

\end{document}